\documentclass[nohyperref]{article}

\usepackage{microtype}
\usepackage{graphicx}
\usepackage{subfigure}
\usepackage{booktabs} 

\usepackage[hyphens]{url}
\usepackage{hyperref}

\usepackage[accepted]{icml2022}

\usepackage{amsmath}
\usepackage{amssymb}
\usepackage{mathtools}
\usepackage{amsthm}

\usepackage[capitalize,noabbrev]{cleveref}

\usepackage{microtype}
\usepackage{url}
\usepackage{amsmath,bm}
\usepackage{mathrsfs}
\usepackage{multirow}
\usepackage{CJKutf8}
\usepackage{verbatim}
\usepackage{amssymb}
\usepackage{caption}
\usepackage{subfigure}
\usepackage{enumitem}
\usepackage{multirow}
\usepackage{makecell}
\usepackage{soul}
\usepackage{arydshln}
\usepackage[bottom]{footmisc}

\newcommand{\tc}{{\mathcal{C}}}
\newcommand{\E}{{\mathbb{E}}}
\newcommand{\KL}{{\mathcal{D}_{\text{KL}}}}
\newcommand{\LL}{{\mathcal{L}}}
\newcommand{\Linput}{\mathcal{L}_{\text{input}}}
\newcommand{\LNAT}{\mathcal{L}_{\text{NAT}}}
\newcommand{\Ltarget}{\mathcal{L}_{\text{target}}}
\newcommand{\Lhattarget}{\hat{\mathcal{L}}_{\text{target}}}
\newcommand{\Lhattargetprime}{\hat{\mathcal{L'}}_{\text{target}}}
\newcommand{\LMPLE}{\mathcal{L}_{\text{MPLE}}}
\newcommand{\hatLMPLE}{\hat{\mathcal{L}}_{\text{MPLE}}}
\newcommand{\Pdata}{P_{\text{data}}}
\newcommand{\Hdata}{H_{\text{data}}}
\newcommand{\Reg}{\mathcal{R}(Q, P_{\text{data}})}
\DeclareMathOperator*{\argmax}{arg\,max}
\DeclareMathOperator*{\argmin}{arg\,min}

\usepackage{pifont}
\newcommand{\smallvspace}{\vspace{0em}}%
\newcommand{\myvspace}[1]{}
\newcommand{\removetext}[1]{}
\newcommand{\klqp}{\KL(Q||P_{\theta})}
\newcommand{\revise}[1]{#1}
\newcommand{\rrevise}[1]{#1}
\newcommand{\rrrevise}[1]{#1}

\theoremstyle{plain}

\theoremstyle{definition}

\theoremstyle{remark}

\icmltitlerunning{On the Learning of Non-Autoregressive Transformers}

\begin{document}

\twocolumn[
\icmltitle{On the Learning of Non-Autoregressive Transformers}

\DeclareUrlCommand\Code{\urlstyle{rm}}
\expandafter\def\expandafter\UrlBreaks\expandafter{\UrlBreaks  
\do\/\do\a\do\b\do\c\do\d\do\e\do\f\do\g\do\h\do\i\do\j\do\k
\do\l\do\m\do\n\do\o\do\p\do\q\do\r\do\s\do\t\do\u\do\v
\do\w\do\x\do\y\do\z
\do\A\do\B\do\C\do\D\do\E\do\F\do\G\do\H\do\I\do\J\do\K
\do\L\do\M\do\N\do\O\do\P\do\Q\do\R\do\S\do\T\do\U\do\V
\do\W\do\X\do\Y\do\Z}



\icmlsetsymbol{equal}{*}
\icmlsetsymbol{special}{\dag}

\begin{icmlauthorlist}
\icmlauthor{Fei Huang}{equal,tsinghua,special}
\icmlauthor{Tianhua Tao}{equal,tsinghua}
\icmlauthor{Hao Zhou}{air,special}
\icmlauthor{Lei Li}{ucsb}
\icmlauthor{Minlie Huang}{tsinghua}
\end{icmlauthorlist}

\icmlaffiliation{tsinghua}{The CoAI group, Tsinghua University. Institute for Artificial Intelligence, State Key Lab of Intelligent Technology and Systems, Beijing National Research Center for Information Science and Technology, DCST, Tsinghua University.}
\icmlaffiliation{air}{Institute for AI Industry Research, Tsinghua University.}
\icmlaffiliation{ucsb}{University of California Santa Barbara}

\icmlcorrespondingauthor{Hao Zhou}{haozhou0806@gmail.com}
\icmlcorrespondingauthor{Minlie Huang}{aihuang@tsinghua.edu.cn}

\icmlkeywords{Non-Autoregressive Text Generation, Machine Translation}

\vskip 0.3in
]



\printAffiliationsAndNotice{\icmlEqualContribution\quad \textsuperscript{\dag}This work is done while Fei Huang was a research intern and Hao Zhou was a research scientist at ByteDance AI Lab.}{} 

\begin{abstract}
Non-autoregressive Transformer (NAT) is a family of text generation models, which aims to reduce the decoding latency by predicting the whole sentences in parallel.
However, such latency reduction sacrifices the ability to capture left-to-right dependencies, thereby making NAT learning very challenging. In this paper, we present theoretical and empirical analyses to reveal the challenges of NAT learning and propose a unified perspective to understand existing successes. First, we show that simply training NAT by maximizing the likelihood can lead to an approximation of marginal distributions but drops all dependencies between tokens, where the dropped information can be measured by the dataset's \textit{conditional total correlation}. Second, we formalize many previous objectives in a unified framework and show that their success can be concluded as maximizing the likelihood on a proxy distribution, leading to a reduced information loss. Empirical studies show that our perspective can explain the phenomena in NAT learning and guide the design of new training methods.
\end{abstract}

\section{Introduction}

\addtolength{\abovedisplayskip}{-0.2em}
\addtolength{\belowdisplayskip}{-0.2em}

\revise{Non-Autoregressive Transformers (NATs, \citealp{nat2018gu}; \citealp{levenshtein2019gu}; \citealp{flowseq2019ma}; \citealp{lexical2021ding}; \citealp{tricktrade2021gu}) have received growing attention due to} \rrevise{their significantly lower decoding latency and approaching accuracy compared to the autoregressive Transformers (ATs) in text generation \cite{glatwmt2021qian, huang2022DATransformer}.
NATs generate the whole sequence in parallel based on the assumption that each token can be predicted independently.}
However, unlike ATs that can be easily trained via Maximum Likelihood Estimation (MLE), NAT learning is very challenging because it drops the left-to-right dependencies.
\citet{nat2018gu} show that directly training NATs via MLE leads to implausible outputs with repeated tokens, revealing their inability to preserve the consistency in generated texts.

To address the problem, many training methods have been proposed.
For example, knowledge distillation (KD, \citealp{seqkd2016kim}; \citealp{nat2018gu}) supervises NATs with target sentences distilled from an AT teacher model.
GLAT \cite{glat2021qian} improves the training by utilizing a masked language model objective.
These methods only change the training objectives without modifying the model, but they demonstrate significant improvements in generation quality.

\definecolor{glatcolor}{rgb}{0.92,0.20,0.14}
\definecolor{spawncolor}{rgb}{0.95,0.66,0.23}
\definecolor{mlecolor}{rgb}{0.92,0.23,0.97}
\begin{figure}[!t]
  \centering
  \includegraphics[width=0.9\linewidth]{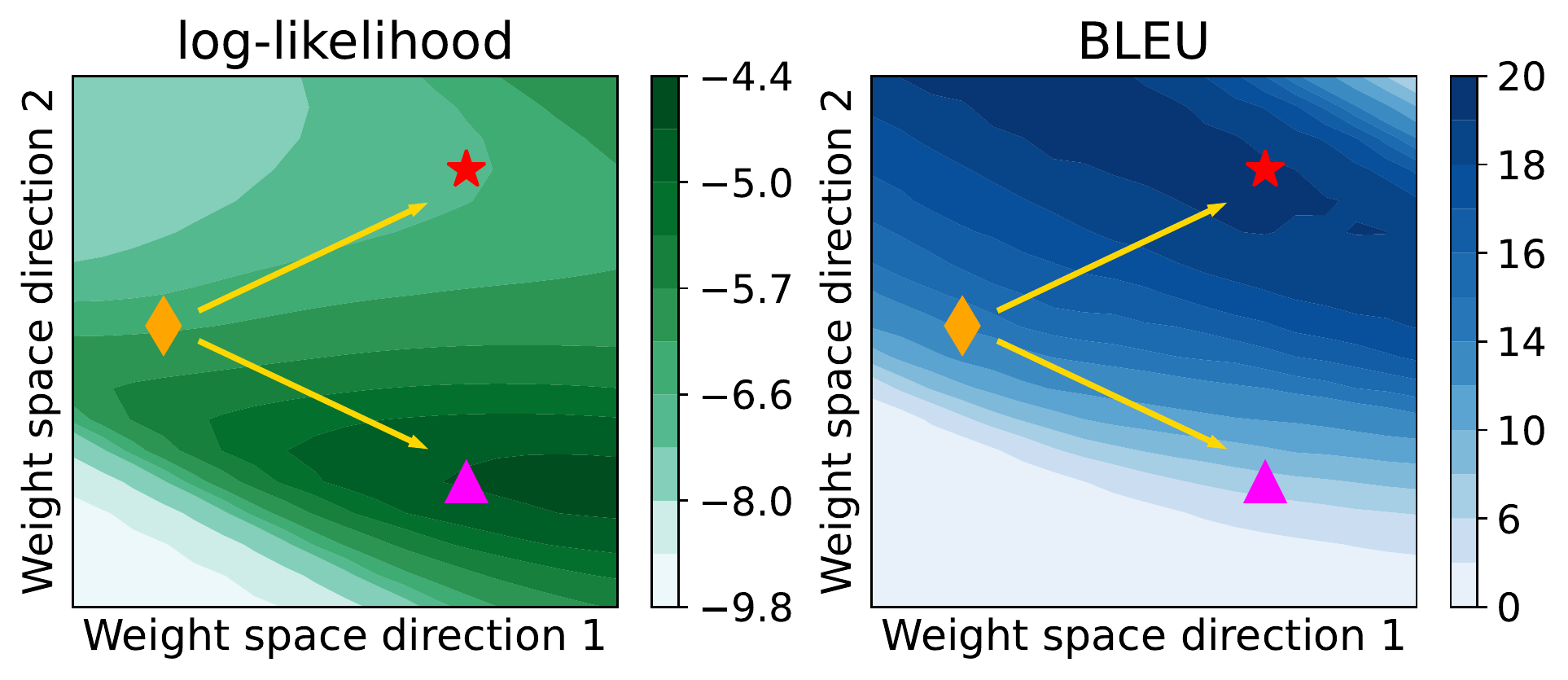}
  \vspace{-0.5em}
  \caption{The log-likelihood and BLEU score on a 2D section in NAT's weight space. GLAT+KD (\textcolor{glatcolor}{$\bigstar$}) and MLE (\textcolor{mlecolor}{$ \blacktriangle $}) are trained via different objectives starting from an initial checkpoint (\textcolor{spawncolor}{$\blacklozenge $}) for 10k steps, \revise{where MLE achieve higher log-likelihood but lower BLEU, and GLAT is the opposite.}
  Each point of the contour map is linearly interpolated from the three checkpoints.
  All values are evaluated on one translation benchmark, the validation set of WMT14 En-De.}
  \label{fig:intro} 
  \vspace{-1.3em}
\end{figure}

\begin{figure*}[!t]
  \centering
  \includegraphics[width=0.9\linewidth]{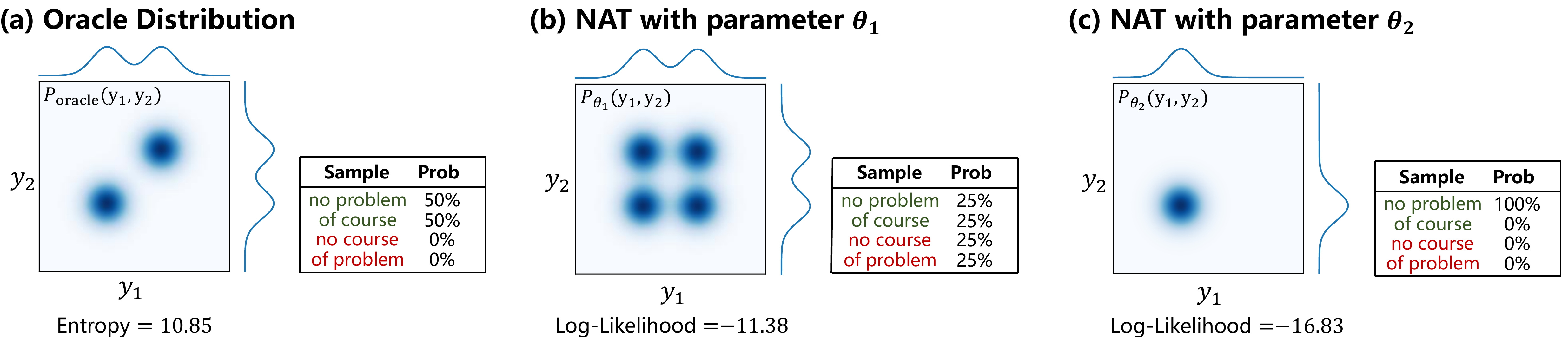}
 \vspace{-0.5em}
  \caption{An example explaining why directly training NATs towards higher likelihood does not lead to better generation quality.
\revise{We illustrate continuous distributions with two variables, i.e., $P(y_1, y_2)$, which are analogous to two-token sentences shown in the table.
Given (a) a two-mode oracle distribution, (b) $P_{\theta_1}$ achieves higher likelihood than (c) $P_{\theta_2}$ but generates undesired outputs that mix the two modes.}
NATs have to follow the independent assumption, i.e., the joint distribution satisfies $P_{\theta}(y_1, y_2) = P_{\theta}(y_1)P_{\theta}(y_2)$.
All distributions are conditioned on a same $X$, which is omitted. }
\label{fig:observations}
\vspace{-1.2em}

\end{figure*}

Despite the empirical successes in NAT learning, there still exists a surprising characteristic not well studied: the objectives leading to a good generation quality actually result in a very low likelihood.\footnote{In this paper, we mainly discuss the NATs that use the same architecture and without iterative refinement unless otherwise specified. The likelihood is obtained on the validation set.}
As shown in Fig.\ref{fig:intro}, we finetune two NATs with different objectives from an initial checkpoint and track the changes in the log-likelihood and the BLEU score.
The optimal training directions under the two metrics are inconsistent, where GLAT+KD improves the generation quality despite that its perplexity is about 10 times of that of the counterpart trained via MLE.

Based on the phenomenons, we raise two questions:
\vspace{-1em}

\begin{itemize}[leftmargin=1em]
    \setlength{\itemsep}{0ex}{
    \setlength{\parskip}{2px}{
    \item \textit{Q1}: Why is NAT learning so challenging that the MLE training does not work well?
    \item \textit{Q2}: Why are previously proposed objectives successful despite they lead to a low likelihood?
    }
    }
\end{itemize}
\vspace{-1em}

In this paper, we present theoretical and empirical analyses to answer the two questions.
For \textit{Q1}, we investigate the challenges of NAT learning from intuitive and theoretical perspectives.
Specifically, we show that directly training NATs towards high likelihood prevents them from learning correct dependencies between target tokens, thereby degrading the generation performance. The lost information can be measured by a property of the data distribution, namely, the \textit{conditional total correlation} \rrevise{(Conditional TC, $\tc$)}, which also measures the difficulties of NAT learning.

For \textit{Q2}, we revisit many previous training objectives and explain their success in a unified framework.
\rrevise{Generally, we find that previous success on NAT can be concluded as adopting a training objective different from the vanilla MLE.
Specifically, instead of maximizing the likelihood on the original dataset $\langle X, Y \rangle$, they in fact approximate a revised distribution, namely, the \textit{proxy distribution} $\langle X+Z, T \rangle$, where $Z$ and $T$ are designed to enhance the inputs and simplify the targets.
With carefully designed $Z$ and $T$, the proxy distribution has a lower $\tc$ than the original distribution, thereby alleviating the information loss in the NAT learning.}
Based on the above analysis, we formulate previous training objectives in a unified framework, named Maximum Proxy-Likelihood Estimation (MPLE). We further derive a general objective to reveal the connections between the proxy distribution and the real distribution, which empirically correlates well with the generation performance and \rrevise{further guides the design of new training methods.}

Our contributions are as follows:
\vspace{-1em}
\begin{itemize}[leftmargin=1em]
    \setlength{\itemsep}{0ex}{
    \setlength{\parskip}{2px}{
    \item We present empirical and theoretical analyses showing that NAT learning is challenging due to the information loss in dependencies, 
    which can be measured by a dataset's property, \rrevise{\textit{conditional total correlation} $\tc$}.
    
    \item We revisit the existing successes in NAT learning and propose to understand previous training objectives in a unified perspective. We reveal that these objectives construct a proxy distribution with \rrevise{a reduced $\tc$}, thereby alleviating the information loss.
    }
    }
\end{itemize}

\section{Challenges of NAT Learning}
\label{sec:broken-mle}

\subsection{Preliminary}

Maximum Likelihood Estimation (MLE) is a widely-used method in training text generation model, which finds a model with the closest distribution to the data distribution in terms of KL divergence \cite{mlekl1998akaike}.
Given a source sentence $X = [ x_1, x_2, \cdots, x_N ]$ and a target sentence $Y = [ y_1, y_2, \cdots, y_M ]$, MLE training minimizes
\begin{align}
    \LL_{\text{MLE}} &= \KL \left[\Pdata(Y|X) || P_{\theta}(Y|X) \right] \label{eq:kl} \\
    &= - \Hdata(Y|X) -\E_{\Pdata(Y|X)}\left[ \log P_{\theta}(Y|X) \right], \notag
\end{align}
where $\Hdata$ is a constant representing the Shannon Entropy, and the second term is the log-likelihood.
For autoregressive Transformers (ATs), the log-likelihood is defined as 
\begin{align}
    \log P_{\theta}^{\text{AT}}(Y|X) = \sum_{i=1}^{M} \log P_{\theta}^{\text{AT}}(y_i|y_{<i}, X), \label{eq:at_ll}
\end{align}
where $y_i$ is predicted based on the prefix $y_{<i}$.

The vanilla NAT makes a \textit{conditional independent assumption} where each token is independent of each other when $X$ is given. Formally, we have
\begin{align}
    \log P_{\theta}^{\text{NAT}}(Y|X) = \sum_{i=1}^{M} \log P_{\theta}^{\text{NAT}}(y_i|X). \label{eq:nat_ll}
\end{align}
Such assumption makes the NAT a poor approximator of the real data distribution, thereby bringing many challenges in NAT learning.
In the next sections, we present an intuitive explanation of the challenges and provide a quantitative method to evaluate the difficulties of NAT learning.

\subsection{Challenges from Intuitive Perspective}
\label{sec:broken-mle-observe}

A major challenge in NAT learning is that directly training NATs towards higher likelihood cannot lead to good generation performance.
We show an intuitive example in Fig.\ref{fig:observations}, which contains
a continuous distribution with two variables as the real distribution (analogous to a sentence with two tokens) and two NATs with different parameters.

\rrevise{Comparing $P_{\theta_1}$ and $P_{\theta_2}$, we find that $P_{\theta_1}$ perfectly approximates the marginal distributions $P(y_1)$ and $P(y_2)$, thereby achieving a higher likelihood.} However, $P_{\theta_1}$ drops the dependency between $y_1$ and $y_2$, leading to wrong outputs by mixing two sentences, previously known as the multi-modality problem \cite{nat2018gu}.
In contrast, although $P_{\theta_2}$ has a low likelihood due to the poor approximations of the marginal distributions, it captures one of the real modes while preserving the correct lexical collocation, i.e., \textit{no} is followed by \textit{problem} but not \textit{course}.

\revise{This example intuitively shows that directly training NAT to maximize the likelihood cannot capture correct lexical collocation due to the severe dependency dropping in target tokens.
In the next section, we quantify the dropped dependencies based on information theory and further evaluate the difficulties of NAT learning.}

\subsection{Challenges from Theoretical Perspective}

\label{sec:total_correlation}

\rrevise{
With the autoregressive decomposition of Eq.\ref{eq:at_ll}, ATs can achieve zero KL divergence theoretically.}\footnote{It is achieved when $P_{\theta}(y_i | y_{<i}, X) = \Pdata(y_i | y_{<i}, X)$.}
\rrevise{However, we show that NATs' KL divergence is bounded by a non-negative constant,} \rrrevise{which corresponds to the information loss in approximating the data distribution.}

\textbf{Theorem 1.} \textit{For a NAT model $P_{\theta}(Y|X)$, we have $\min_{\theta} \KL[\Pdata(Y|X) || P_{\theta}(Y|X)] \geq \tc$, where $\tc = \sum_{i=1}^{M} \Hdata(y_i|X) - \Hdata(Y|X)$, and $\Hdata(\cdot|X)$ is the Shannon Entropy}.
\vspace{-1.5em}

{
\footnotesize
\begin{align}
    & \textit{Proof.}\quad \KL[\Pdata(Y|X) || P_{\theta}(Y|X)] \notag \\
    & = - \Hdata(Y|X) - \E_{\Pdata(Y|X)} \left[ \sum_{i=1}^{M} \log P_{\theta}(y_i|X) \right]   \notag \\
    & \quad\quad\quad\quad\quad\text{\textit{(Conditional Independent Assumption of Eq.\ref{eq:nat_ll})}}\notag \\
    & = - \Hdata(Y|X) - \sum_{i=1}^{M} \E_{\Pdata(y_i|X)} \left[ \log P_{\theta}(y_i|X) \right]   \notag \\
    & \geq - \Hdata(Y|X) + \sum_{i=1}^{M} \Hdata(y_i|X) \quad\text{\textit{(Gibbs' Inequality)}} \notag
\end{align}
}
\vspace{-1.5em}

\noindent \revise{The equality is achieved when $P_{\theta}(y_i|X) = \Pdata(y_i|X)$.}
Note that $\tc$ is a non-negative constant called \textit{conditional total correlation} (Conditional TC, \citealp{totalcorrelation1960watanabe}) or \textit{multi-information} \cite{multiinfomration1998studeny}, which measures the information of dependencies between the target tokens when $X$ is known.
\revise{We make two remarks on Theorem 1:}

\revise{\textbf{Remark 1.} \textit{A well-trained NAT (in terms of KL divergence) achieves perfect approximations on marginal distributions but drops all the dependency information between target tokens, which can be measured by $\tc$.}}

\revise{\textbf{Remark 2.} \textit{$\tc$ is a property of data distribution representing the difficulties in NAT learning. Given the data distribution, an NAT cannot achieve an information loss less than $\tc$ regardless of its parameters or training methods.}}

\begin{table}[t]
\caption{Estimated $\tc$ and the gap of BLEU between AT and NAT on various datasets.\footnotemark\ \revise{A large $\tc$ leads to significant performance gap between AT and NAT.} All models are trained via MLE. $\Delta\text{BLEU} = \text{BLEU}_{\text{AT}} - \text{BLEU}_{\text{NAT}}$.
The dataset details are presented in Appendix \ref{app:cgap}.}
\vspace{-0.8em}
\begin{center}
\resizebox{0.9\linewidth}{!}{
\setlength{\tabcolsep}{2mm}{
\begin{tabular}{c|cccc}
\hline
\bf Dataset         & \bf $\tc$ & \bf $\Delta$BLEU & \bf $\text{BLEU}_{\text{AT}}$ & \bf $\text{BLEU}_{\text{NAT}}$ \\
\hline
WMT14 En-De & 2.50 & 15.32 & 27.11 & 11.79 \\
WMT16 En-Ro & 2.20 & 9.98 & 33.70 & 23.72 \\
Synthetic B & 1.51 & 5.66 & 20.97 & 15.31 \\
Synthetic A & 0.92 & 0.35 & 26.96 & 26.61 \\
\hline
\end{tabular}
}
}
\end{center}

\label{tab:total-correlation}
\vspace{-1.5em}

\end{table}
\footnotetext{We report $\tc$ with the \textit{V-entropy} \cite{vinformation2020xu} instead of the Shannon entropy because the latter is intractable due to the unknown data distribution. \revise{Note that $\tc$ measures the amount of information, whose unit is \textit{bit}, which is comparable across datasets. However, $\Delta$BLEU is not strictly comparable, where we present a more rigorous comparison in Appendix \ref{app:cgap}.}}

\begin{figure*}[!t]
\begin{minipage}[b]{.60\textwidth}
\centering
\includegraphics[width=0.85\textwidth]{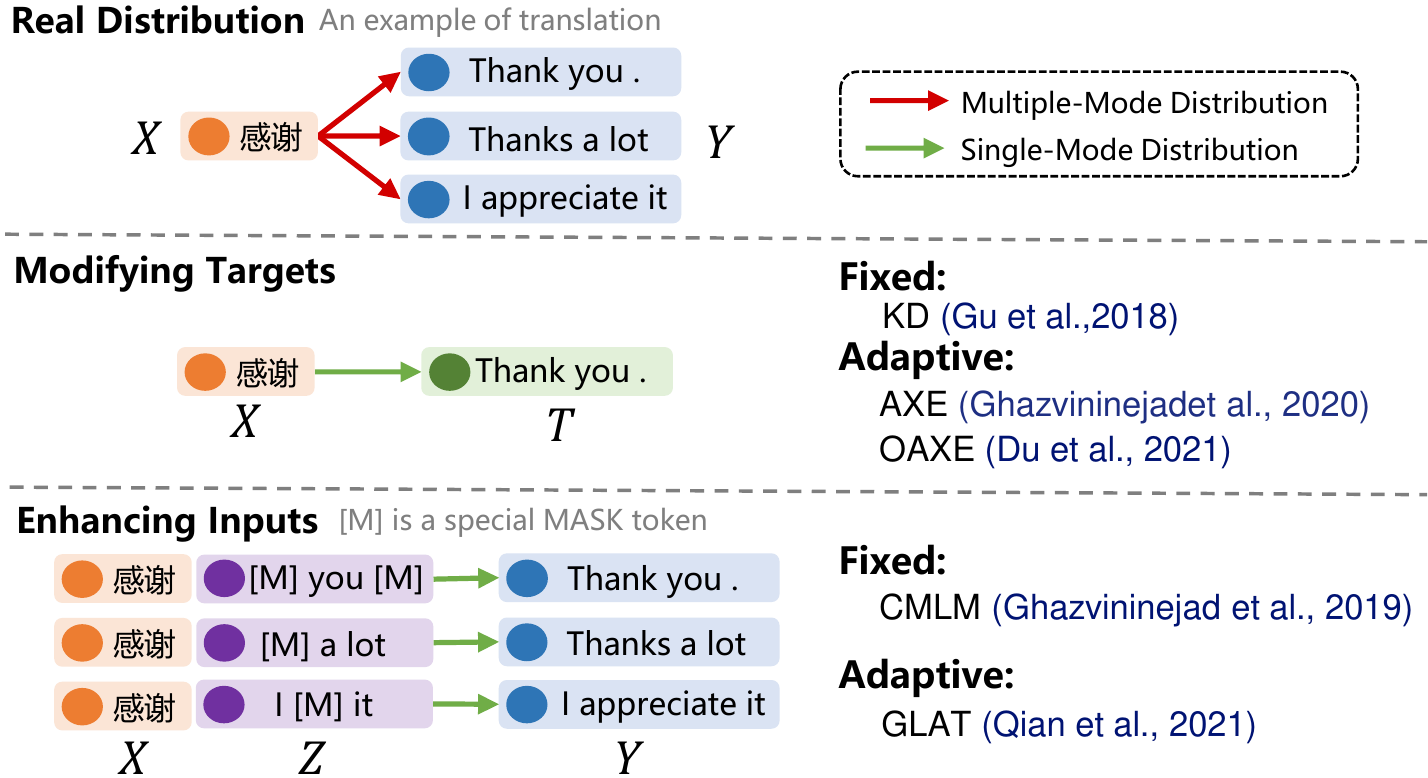}
\vspace{-0.5em}
\caption{An overview of methods to construct the proxy distribution $Q$, which fall into two categories: \textit{Modifying Targets} (replacing $Y$ by a proxy target $T$) and \textit{Enhancing Inputs} (with a proxy input $Z$). \textit{Fixed} and \textit{Adaptive} indicate whether the proxy distribution is adjusted through the training.}
\label{fig:model}
\end{minipage}
\hfill
\begin{minipage}[b]{.37\textwidth}
\centering
\includegraphics[width=0.96\textwidth]{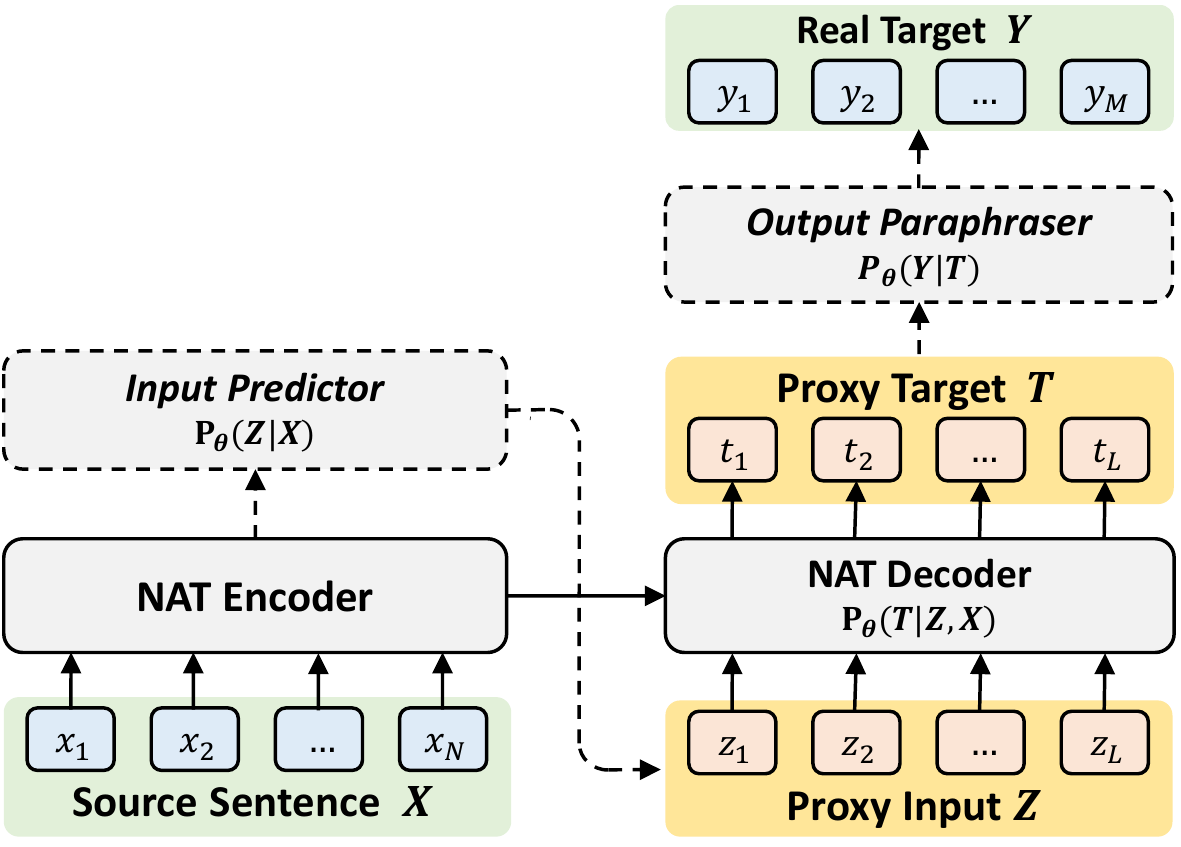}
\vspace{-0em}
\caption{The latent variable model used in the derivation. The source sentence $X$ and the real target $Y$ are observable, whereas the proxy input $Z$ and the proxy target $T$ are latent.}
 \label{fig:architecture}
\end{minipage}
\vspace{-1em}
\end{figure*}


\textbf{Conditional TC and Performance Gap}\ \ \ 
To better understand how $\tc$ affects the NAT performance,
we estimate $\tc$ and compare the generation performance of AT and NAT models trained via MLE on four datasets.
\rrrevise{Since $\tc$ is usually high for most datasets,}
\rrevise{besides two translation benchmarks, we further construct two synthetic datasets that have a lower $\tc$.}
Please refer to Appendix \ref{app:cgap} for more details.

\revise{As shown in Table \ref{tab:total-correlation}, large $\tc$ indicates strong dependencies between target tokens, leading to a serious performance gap between NAT and AT models.
When $\tc$ is small, NAT can achieve a similar performance with AT, verifying that the large Conditional TC is the main obstacle in NAT learning.}

\section{Understanding NAT Learning via Maximum Proxy-Likelihood Estimation}

\label{sec:MPLE}

Sec.\ref{sec:broken-mle} shows that MLE-trained NAT drops the dependencies between tokens, where $\tc$ measures the difficulties in NAT learning.
In this section, we investigate previous successes in training NATs and propose a unified perspective to understand them.

Specifically, we revisit existing training objectives and find that many of them improve the MLE training by simplifying the target sentences or enhancing the training inputs.
\revise{Such modifications significantly change the training directions, where they actually maximize the likelihood on a \textit{proxy distribution} instead of the original distribution.
The proxy distribution with modified targets or inputs usually has a low $\tc$, thereby reducing the information loss in NAT learning. }
Based on the above analysis, we formulate these methods in a unified framework, named Maximum Proxy-Likelihood Estimation~(MPLE).
Intuitively, MPLE's objective can be expressed as
\begin{gather}
    \LL = \klqp + \Reg. \label{eq:preview}
\end{gather}
The first term is similar to the MLE objective, which trains the model towards the proxy distribution $Q$ instead of $\Pdata$. The second term is a regularizer controlling the distortion between $Q$ and $\Pdata$.

\subsection{Revisiting Previous Successes}
\label{sec:revisit}

Considering the severe challenges in NAT learning, many training methods are proposed to improve the generation performance.
For example, \textit{Aligned Cross Entropy} (AXE, \citealp{axe2020ghazvininejad}) finds that the cross-entropy loss highly penalizes small shifts in word order, which deviates from the evaluation of generation quality and thus hinders the NAT training. They propose an aligned-based objective that allows small target shifts to alleviate the problem.
\textit{GLAT} \cite{glat2021qian} proposes to promote representation learning by utilizing curriculum learning. Specifically, they train NATs similar to the masked language model, which feeds a masked target as the decoder input and adjusts the training difficulties by annealing the masking ratio.

Although these methods are proposed with different motivations, we find that they generally share a similar objective that can be interpreted as the MLE training.
Specifically, they still use the cross-entropy loss between the NAT predictions and the target tokens, except that the target labels or model inputs are changed.\footnote{They also do not change the inference process, e.g., no extra inputs are introduced for decoder in generation.}
Then we can interpret the loss as an objective of MLE, but the target distribution is actually replaced by a new distribution with their new inputs and outputs, where we call it a \textit{proxy distribution} $Q$.
In these methods, NATs are trained on the proxy distribution $Q$ to maximize the likelihood, which explains why they have a low likelihood on the original validation set.

By examining these methods closely, we find that the proxy distribution $Q$ is an essential key to their success.
Specifically, we divide existing methods into two categories: \rrevise{\textit{Modifying Targets} or \textit{Enhancing Inputs}}.
As shown in Fig.\ref{fig:model}, both approaches try to preserve a one-to-one mapping between the new inputs and outputs, which intuitively reduces $\tc$ by limiting the possible modes in the proxy distribution,\footnote{A distribution with multiple modes requires dependency information to recover the joint distribution, as shown in Fig.\ref{fig:observations}.}
thereby alleviating the information loss in NAT learning.

Formally, we denote the proxy distribution by $Q(T|Z, X)$, where the original $Y$ is replaced by a \textit{proxy target} $T$, and the original $X$ is enhanced with a \textit{proxy input} $Z$. 
Next, we revisit existing methods of NAT learning to study how they construct the proxy distribution $Q$. 

\rrevise{\textbf{Constructing $Q$ by Modifying Targets ($Y \rightarrow T$)}}\ \  
Seq\-uence-level knowledge distillation (KD, \citealp{nat2018gu}) is a direct method to simplify the targets. For a given input $X$, an autoregressive teacher generates the proxy target $T$ by beam search, which replaces the diverse references and thus reduces the possible outputs in the data distribution.
The KD data are usually generated in advance and does not change during NAT training.

Some methods construct $T$ through the training, which are adaptively adjusted according to the NAT model. AXE~\cite{axe2020ghazvininejad} and OaXE~\cite{oaxe2021du} use alignment-based objectives, which match each prediction with a reference token and calculate the cross-entropy loss.
The two losses are equivalent to obtaining the MLE objective with a new target $T$, where $T$ is a permutation of $Y$ but closer to the model prediction.

\rrevise{\textbf{Constructing $Q$ by Enhancing Inputs ($X\rightarrow Z, X$)}}\ \ \ \ 
CMLM\footnote{We discuss the non-iterative version of CMLM here, following \citet{axe2020ghazvininejad, oaxe2021du}.} \cite{cmlm2019ghazvininejad} uses a masked language model objective, where a randomly masked target sentence is fed into the NAT decoder.
Intuitively, if $P(Y|X)$ has multiple possible outputs, $Q(Y|Z, X)$ can reduce the number of candidates with the constraint of $Z$, which again leads to a simplified distribution with reduced $\tc$.

Unlike CMLM that samples $Z$ from a predefined distribution by random masking, GLAT~\cite{glat2021qian} proposes to sample $Z$ adaptively according to the NAT performance. \revise{Specifically, if the NAT well approximates $P(Y|X)$ without $Z$, GLAT will mask most tokens in the proxy input. Since the NAT uses full masks in inference, GLAT im\-proves CMLM by reducing the training and inference gap.}

\subsection{A Unified Objective of MPLE}
\label{sec:deriv_MPLE}

Existing methods simply train NAT by maximizing the likelihood on the proxy distribution $Q(T|Z,X)$. 
However, they do not answer \textit{when the performance on the proxy distribution can generalize to the real distribution.} 
For example, a good approximation of $Q$ may not guarantee good generation performance on $\Pdata$ since there can be a substantial distortion between the two distributions.

In MPLE, we propose considering $\tc$ and the data distortion together in a unified objective.
Specifically, we regard $T$ and $Z$ as latent variables and build a latent variable model that connects $Z, T$ and $X, Y$, as shown in Fig.\ref{fig:architecture}. Formally,
\vspace{-1.2em}

{
\footnotesize
\begin{gather}
    P_{\theta}(Y|X) = \sum_Z \sum_T P_{\theta}(Y|T) P_{\theta}(T|Z, X) P_{\theta}(Z|X), \label{eq:latent}
\end{gather}
}

\vspace{-0.8em}
where $P_{\theta}(T|Z, X)$ is the NAT decoder, and the other two modules bridge $Z$ with $X$, and $T$ with $Y$, respectively.
Then we derive our objective from the likelihood on $\Pdata(Y|X)$:
\vspace{-1.5em}

{
\footnotesize
\begin{align}
    &\ \ \ \ - \E_{\Pdata(Y|X)}\log P_{\theta}(Y|X) \notag \\
    &= -\E_{\Pdata(Y|X)} \log \left[ \E_{Q(T, Z|X)} \frac{P_{\theta}(Y, T, Z|X)}{Q(T, Z|X)} \right] \notag \\ 
    & \leq -\E_{\Pdata(Y|X)} \E_{Q(T, Z|X)} \left[ \log \frac{P_{\theta}(Y, T, Z|X)}{Q(T, Z|X)} \right] \label{eq:variational} \\ 
    & = -\E_{\Pdata(Y|X)} \E_{Q(T, Z|X)} \Big[ \log P_{\theta}(Y|T) + \notag \\
    & \quad\quad\quad\quad\quad\quad\quad \log \frac{P_{\theta}(T|Z, X)}{Q(T|Z, X)} + \log \frac{P_{\theta}(Z|X)}{Q(Z|X)} \Big] \label{eq:complex_terms}
\end{align}
}

\vspace{-0.5em}
In Eq.\ref{eq:variational}, we apply \textit{variational principle} \cite{variaionaltutorial2012fox} by introducing $Q(T, Z|X)$, which specifies how we obtain $Z, T$ and
can be decomposed into the proxy distribution $Q(T|Z, X)$ and $Q(Z|X)$. 

Eq.\ref{eq:complex_terms} is our unified objective $\LMPLE$, which can be simplified and recovers our intuition in Eq.\ref{eq:preview}:
\vspace{-1.5em}

{
\footnotesize
\allowdisplaybreaks
\begin{align}
    \LMPLE &= \underbrace{\LNAT}_{\klqp} + \underbrace{\Ltarget + \Linput}_{\Reg}, \label{eq:objective} \\
    \LNAT &= \E_{Q(Z|X)}\KL \left[ Q(T|Z, X) || P_{\theta}(T|Z, X) \right], \label{eq:L_nat} \\
    \Ltarget &= \E_{\Pdata(Y|X)} \E_{Q(T|X)} \left[ - \log P_{\theta}(Y|T) \right], \label{eq:L_target} \\
    \Linput &= \KL \left[ Q(Z|X) || P_{\theta}(Z|X) \right].  \label{eq:L_input}
\end{align}
}

\vspace{-0.5em}
In Eq.\ref{eq:objective}, $\LNAT$ supervises the decoder $P_{\theta}(T|Z, X)$ to maximize the likelihood on the proxy distribution.
$\Ltarget$ and $\Linput$ measure the cost in bridging $T, Z$ with $X, Y$, and act as regularizers to avoid large distortions between the proxy and original variables.

\revise{Moreover, since $P_{\theta}(T|Z, X)$ still follows the independent assumption, we can derive a lower bound of $\LNAT$ in a similar way of Theorem 1. Specifically, we have}
\vspace{-1.5em}

{
\footnotesize
\begin{align}
\textstyle \tc' := \E_{Q(Z|X)} \left[ \sum\nolimits_{i=1}^{L} H_Q(t_i|Z, X) - H_Q(T|Z, X) \right], \label{eq:c'}
\end{align}
}

\vspace{-1em}
\revise{that satisfies $\LNAT \geq \tc'$, where $\tc'$ is the Conditional TC of the proxy distribution $Q(T|Z, X)$.}

\subsection{Understanding Existing Methods in MPLE}
\label{sec:understanding}

$\LMPLE$ seems a bit complex because it includes both the likelihood term to train the NAT model and the objective for selecting the proxy inputs and targets.
To understand previous methods in MPLE, we describe the training process as an \textit{Expectation Maximization} algorithm including two steps:
\revise{(1) find optimal proxy distribution $Q$ by adjusting the proxy variables $Z$ and $T$; (2) optimize model parameter $\theta$.}

In \textbf{E-step}, \rrevise{we fix the model parameter $\theta$ and update proxy variables to reduce $\LMPLE$, which aims to find good proxy distribution $Q$ to balance $\LNAT$ and the data distortion.
Since $\theta$ is fixed, adjusting proxy variables for lower $\LNAT$ does not affect the NAT model but actually optimizes $\tc'$, where $\LNAT$ is the upper bound of $\tc'$ defined in Eq.\ref{eq:c'}.}

However, such optimization is non-trivial, where existing methods utilize some heuristic rules.
For example, \rrrevise{KD obtains $T$ by distilling sentences from a pre-trained AT teacher, which efficiently alleviates the information loss by reducing the modes in the dataset.}
\rrevise{AXE and OaXE obtain $T$ by aligning the NAT prediction with $Y$, where they have hyper-parameters for controlling the distortion within an acceptable range.}
Please refer to Appendix \ref{app:instances} for more details about the heuristic rules in existing methods.

\revise{As introduced in Sec.\ref{sec:revisit}, these heuristic rules utilize either fixed or adaptive strategies.}
Fixed strategies obtain the proxy distribution before the training, where $\tc'$ is lower than the original $\tc$, but not further optimized.
In contrast, adaptive strategies adjust the proxy distribution through the training, which usually outperforms the fixed ones.

In \textbf{M-step}, we fix the proxy distribution $Q$ and optimize the model $\theta$. Since all $Q$'s entropies are constants and thus ignored, the three losses in Eq.\ref{eq:objective} can be easily calculated based on $Z$ and $T$ previously obtained in the E-step.
Specially, $\LNAT$ recovers the objectives of existing methods by maximizing the likelihood on the proxy distribution.

\textbf{Quantifying Data Distortion}\ \ \ 
Existing methods heuristically obtain $Z$ and $T$ to balance the training difficulties and the data distortion, which does not involve a measurement of the distortion.
MPLE provides a method to quantify the distortion, allowing for comparisons between different methods in constructing the proxy distribution.

Specifically, we use the output paraphraser $P_{\theta}(Y|T)$ and the input predictor $P_{\theta}(Z|X)$ to measure the data distortion $\Ltarget$ and $\Linput$, respectively.
For $\Ltarget$, we define the output paraphraser as a simple non-trainable distribution related to the similarity between $Y$ and $T$:
\begin{align}
P_{\theta}(Y|T) = exp(\beta S(Y, T)) / \zeta, \label{eq:bleu-distribution}
\end{align}
where $\beta$ is a hyper-parameter, $S(Y, T)$ is the sentence BLEU, and $\zeta = \sum_{Y} exp(\beta S(Y, T))$. However, the normalization term $\zeta$ is intractable, so we drop it and empirically use $\Lhattarget$ instead:
\begin{align}
    \Lhattarget &= \E_{\Pdata(Y|X)} \E_{Q(T|X)} \left[ -\beta S(Y, T) \right]. \label{eq:L_target_hat}
\end{align}
Intuitively, Eq.\ref{eq:L_target_hat} measures the distortion between proxy and real targets by the average BLEU score.

For $\Linput$, we design a trainable input predictor specially for GLAT and CMLM, where the other methods without an extra input $Z$ always have $\Linput=0$.
Specifically, we define the input predictor as a classifier, which predicts $z_i$ from the vocabulary including a special mask token. We predict $Z$ non-autoregressively (See Appendix \ref{app:cmlm_glat} for details):
{
\begin{align}
    \log P_{\theta}(Z|X) &= \sum_{i=1}^{M} \log P_{\theta}(z_i|X). \label{eq:input_predictor}
\end{align}
}
Then, $\Linput$ can be calculated according to Eq.\ref{eq:L_input}.

\rrevise{\textbf{Discussing More Work from MPLE Perspective}\ \ \ \ 
Besides the methods discussed above, MPLE can also explain many other objectives proposed for NAT learning, including (1) the methods introducing continuous or discrete latent variables~\cite{discretelatent2018kaiser, vae2020shu, CNAT2021bao, latentglat2022}; (2) enhancing NAT decoder with order information \cite{pnat2019bao, reorder2021ran}, POS taggings \cite{posconstrain2021yang}, or tokens sampled from target sentences \cite{dslp2021huang}; (3) KD variants like reverse distillation \cite{reversekd2021ding} or repeated distillation \cite{kdnat2020zhou, emnatsun2020}.}

\rrevise{Notably, CTC-based methods~\cite{ctc2018libovicky, imputermt2020saharia} and DA-Transformer~\cite{huang2022DATransformer} have been shown very effective in NAT learning, where they also utilize alignment-based objectives. 
Unlike AXE, these methods predict a sequence longer than the real target, and then remove useless tokens by rules or model predictions.
\rrrevise{Their success show that the proxy target $T$ does not necessarily have similar length with $Y$, where a longer $T$ can be more flexible and efficient in reducing the token dependencies. Moreover, they introduce a different $P_\theta(Y|T)$ from Eq.\ref{eq:bleu-distribution}, which predicts $Y$ from a longer $T$ with reconstruction of dependency information, e.g., by transitions predicted in DA-Transformer.}\footnote{\rrevise{Both methods do not directly fit in Eq.\ref{eq:objective} because they maximize the logarithm of probability sum on all alignments instead of a single $T$. However, we refer the reader to Sec.3.2 of \citet{huang2022DATransformer}, which shows that their objectives can be regarded as utilizing multiple proxy targets with different weights.}}}

\rrevise{Finally, MPLE also connects with iterative NATs \cite{iterativerefinement2018lee, cmlm2019ghazvininejad, disco2020kasai, jmnat2020guo}. Although iterative NATs do not satisfy the independent assumption in Eq.\ref{eq:nat_ll}, they still predict tokens independently in each iterative step. Specifically, we point out that (1) $\tc$ measures the information loss of iterative NAT in each refinement step; (2) some iterative NATs are special cases of MPLE with parameter sharing in Input Predictor and NAT decoder. Please refer to Appendix \ref{app:iterative} for details.
}

\subsection{A KD Variant from MPLE}
\label{sec:dynamic-kd}

Existing methods heuristically obtain $Z$ and $T$ to construct the proxy distribution.
We propose a new variant of KD that improves the proxy distribution by explicitly balancing $\LNAT$ and the data distortion, named dynamic KD.

For a source sentence $X$, we obtain a target candidate set $\Gamma$, which contains the raw data and distilled data from AT teachers of different sizes, i.e., Transformer-\textit{tiny/small/base/big}.
Then we choose a best target $T \in \Gamma$ that minimizes $\LNAT + \Lhattarget$.
Noticing that Eq.\ref{eq:L_target_hat} is intractable due to the sampling from $\Pdata$, we use the pairwise BLEU between the candidates instead. More details are presented in Appendix \ref{app:dynamic_kd}.

Previous work \cite{kdnat2020zhou} finds that the KD data from a larger AT teacher is closer to the real data but more difficult to predict, where they suggest choosing the teacher size according to NAT's capacity. 
Our method dynamically selects the best proxy target from multiple KD candidates for each sample, which achieves substantial improvement over NATs trained on any single KD data.

\section{Experiments}

\label{sec:experiment_setting}

\textbf{Dataset}\ We use two translation benchmarks, WMT14 En-De (4.5M) and WMT17 Zh-En (20M), and follow \citet{kdnat2020zhou, disco2020kasai} for preprocessing.

\smallvspace
\noindent \textbf{Knowlegde Distillation}\ \  We use Transformer-\textit{base} with the same settings as \citet{transformer2017vaswani} and generate the distilled data with beam size 5. All models are based on KD unless otherwise specified.

\smallvspace
\noindent \textbf{Implementation Details}\ \ We implement Raw Data, KD, AXE, OaXE for obtaining proxy targets, and Vanilla (no extra input), CMLM, GLAT for obtaining proxy inputs.
We generally follow the hyper-parameters in \citet{glat2021qian}.
For fair comparisons, we only modify the heuristic rules to obtain $Z$ and $T$, which may be different from their original implementations. For example, we do not use iterative refinement for CMLM, or
combine OaXE with CMLM.
Unless otherwise specified, we do not utilize reranking methods or other decoding tricks.
More details are in Appendix \ref{sec:experiment-details}.

\smallvspace
\noindent \textbf{Metrics}\ \  
We utilize tokenized BLEU \cite{bleu2002papineni} to evaluate the translation performance.
$\Linput$ and $\LNAT$ are averaged per token on validation set. $\Lhattarget$ in Eq.\ref{eq:L_target_hat} requires multiple real targets $Y$ from $\Pdata$, so we utilize multi-reference annotations from \citet{uncertainty2018ott,parity2018hassan}, where each sample has 10(2) extra human-annotated references for En-De(Zh-En). To calculate $\Lhattarget$ by Eq.\ref{eq:L_target_hat}, we use $\beta=0.2$ for En-De, $\beta=0.25$ for Zh-En. $\hatLMPLE := \LNAT+\Linput+\Lhattarget$. We measure the speedup of decoding latency on WMT14 En-De with batch size 1.

\begin{figure}[!b]
  \centering
  \vspace{-2em}
  \includegraphics[width=0.75\linewidth]{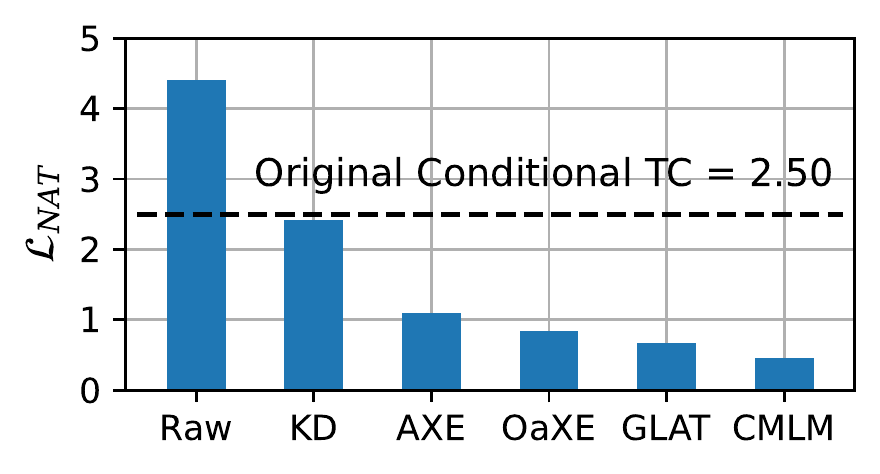}
  \vspace{-1em}
  \caption{\revise{$\LNAT$ of different methods on WMT14 En-De. All methods except Raw Data achieve lower $\LNAT$ than the dataset's $\tc$, verifying that training on the proxy distribution alleviates the information loss.}}
  \label{fig:limitation_break}
\end{figure}

\subsection{Verification of Reduced Information Loss}
\label{sec:exp_limit_break}

\revise{Theorem 1 implies that any NAT approximating the real distribution cannot achieve less information loss than dataset’s $\tc$. We argue that existing methods approximate a proxy distribution $Q$ instead, thereby achieving reduced information loss.
To verify the proposition, we compare $\LNAT$ of different methods against the the original dataset’s $\tc$. }

\revise{As shown in Fig.\ref{fig:limitation_break}, most methods except Raw Data achieve lower $\LNAT$ than the dataset's $\tc$.
Note that $\LNAT$ evaluates the information loss in approximating the proxy distribution, which is the upper bound of $Q$'s Conditional TC, i.e., $\tc'$ defined Eq.\ref{eq:c'}.}
The results empirically verify that (1) training on the proxy distribution alleviates the information loss in NAT learning; (2) the proxy distribution has a reduced Conditional TC.

However, lower $\LNAT$ does not promise higher BLEU because they do not control the data distortion.
In the next sections, we will analyze how different methods affect the performance by balancing $\LNAT$ and the data distortion.

\subsection{Effects of Proxy Target}
\label{sec:analysis_target}

\begin{table}[!t]
\caption{Comparison of proxy targets on WMT14 En-De. All methods use Vanilla for $Z$ and $\Linput=0$. $\hatLMPLE$ and BLEU are strongly correlated (Pearson's $|r|$=0.99). AXE's $\tau$ and OaXE's numbers indicate the skip penalty and pre-training steps, which are hyper-parameters in choosing $T$.}
\vspace{-0.5em}
\begin{center}
\resizebox{0.75\linewidth}{!}{
\setlength{\tabcolsep}{1.0mm}{
\begin{tabular}{l|cccc}
\hline
\bf Models & \bf $\LNAT$ & \bf $\Lhattarget$ & \bf $\hatLMPLE$ & \bf BLEU \\
\hline
Raw Data & 4.41 & -6.42 & -2.01 & 11.79 \\
KD & 2.42 & \bf -7.08 & -4.66 & 20.87 \\
\hline
+ AXE ($\tau$=1) & \bf 0.78 & -5.13 & -4.35 & 18.56 \\
+ AXE ($\tau$=5) & 1.09 & -6.34 & -5.25 & 22.22 \\
+ AXE ($\tau$=10) & 1.25 & -6.50 & -5.26 & 22.35 \\
\hline
+ OaXE (10k) & 1.03 & -4.41 & -3.38 & 15.00 \\
+ OaXE (50k) & 0.79 & -5.84 & -5.06 & 21.37 \\
+ OaXE (300k) & 0.83 & -6.28 & \bf -5.44 & \bf 22.76 \\
\hline
\end{tabular}
}
}
\end{center}
\label{tab:compare_target}
\vspace{-1.5em}
\end{table}

\begin{figure}[!b]
  \centering
  \vspace{-1em}
  \includegraphics[width=0.65\linewidth]{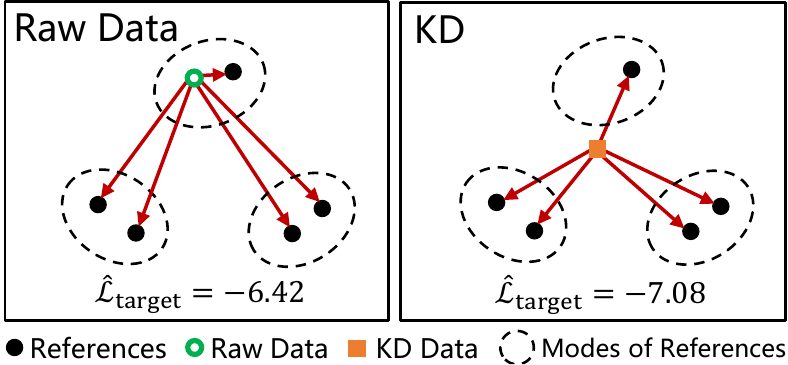}
  \vspace{-0.2em}
  \caption{\revise{KD data are even closer to the multiple references on average than Raw Data on WMT14 En-De, thereby achieving lower $\Lhattarget$ and improving the performance.}}
  \label{fig:kd_raw}
  \end{figure}

In this section, we compare different methods of obtaining proxy targets with varying hyper-parameters. We present the results on En-De in Table \ref{tab:compare_target} and Zh-En in Appendix \ref{app:results_on_zhen}.

\smallvspace
\noindent \textbf{Strong Correlation.} \ \ 
$\hatLMPLE$ is strongly correlated with BLEU, where $\LNAT$ and $\Lhattarget$ are both important. 
For example, AXE($\tau$=1) achieves low $\LNAT$ with high $\Lhattarget$, indicating that $T$ is easy to predict but heavily distorted from the real target. On the contrary, KD's proxy target is less distorted but hard to predict.
OaXE(300k) balances the two losses well and thus achieves the best BLEU.

$\beta$ in Eq.\ref{eq:L_target_hat} will affect the scale of $\Lhattarget$, where we choose $\beta=0.2$ to maximize the correlation. However, the choose of $\beta$ is not sensitive that $|r| \geq 0.8$ for all $\beta \in [0.1, 0.5]$.

\smallvspace
\noindent \textbf{Secret Advantage of KD.} \ \ 
Previous work \cite{nat2018gu} has shown that KD can simplify the training data and thus reduce $\LNAT$.
\revise{However, our results \rrevise{show} a secret advantage that KD also achieves the lowest $\Lhattarget$, indicating that the KD data are even closer to the multiple human references on average than the raw data.
This result is caused by the diversity of human annotations, as shown in Fig.\ref{fig:kd_raw}.} Although the KD data may not belong to any modes of the data distribution, it still has higher similarity on average.

\smallvspace
\noindent \textbf{Hyper-parameters and Trade-off.}\ \ 
AXE and OaXE utilize tricks to avoid large distortion between the proxy target and the real target. For example, AXE tunes the skip penalty $\tau$, and OaXE tunes the pre-training step. MPLE provides a quantifiable method to measure the trade-off between the likelihood loss $\LNAT$ and the distortion $\Lhattarget$, which improves the interpretability of hyper-parameter selection.

\subsection{Effects of Proxy Input}
\label{sec:analysis_latent_input}

We compare methods that obtain proxy inputs including several variants of CMLM and GLAT, which are also used in \citet{glat2021qian}. These variants use different strategies for masking, whose details are presented in Appendix \ref{app:cmlm_glat}.

In the inference of CMLM and GLAT, they use a full mask as the proxy input by default (Default Decoding), leading to a large gap between train and inference.
\revise{We propose to sample the latent input (Input Sampling) based our latent variable model in Eq.\ref{eq:latent}: We first sample $Z$ according to the input predictor $P_{\theta}(Z|X)$,\footnote{More precisely, we first decide whether a token is masked according to the predicted distribution. If it is not masked, we directly use the most likely non-mask token, which empirically leads to better performance. Please see Appendix \ref{app:cmlm_glat} for details.} and then choose the most likely tokens predicted by the NAT decoder.}
We present the results on En-De in Table \ref{tab:compare_input} and Zh-En in Appendix \ref{app:results_on_zhen}.

\begin{table}[!t]
\caption{Comparison of proxy inputs on WMT14 En-De. All methods use KD for $T$ and $\Lhattarget=-7.08$. Sample and Default indicate the BLEU score with Input Sampling and Default Decoding. $\hatLMPLE$ is strongly correlated with Sample BLEU ($|r|$=0.99) but less correlated with Default BLEU ($|r|$=0.37). \revise{The variants of CMLM and GLAT use different strategies for masking, detailed in Appendix \ref{app:cmlm_glat}.}}
\vspace{-0.5em}
\begin{center}
\resizebox{0.9\linewidth}{!}{
\setlength{\tabcolsep}{1.0mm}{
\begin{tabular}{l|ccccc}
\hline
\bf Models & \bf $\Linput$ & \bf $\LNAT$ & \bf $\hatLMPLE$ & \bf Sample & \bf Default \\
\hline
Vanilla & \bf 0 & 2.42 & -4.66 & 20.87 & 20.87\\
\hline
CMLM & 0.99 & \bf 0.46 & -5.63 & 23.48 & 19.39\\
+ fixed masking ratio & 0.48 & 0.55 & \bf -6.05 & 24.28 & 19.35\\
\hline
GLAT & 0.45 & 0.66 & -5.96 & 23.98 & 25.12\\
+ Levenshtein dist. & 0.41 & 0.73 & -5.94 & 24.03 & 24.84\\
+ mask by $P_{\text{ref}}$ & 0.25 & 1.24 & -5.59 & 22.98 & 24.22\\
+ mask by $1-P_{\text{ref}}$ & 0.57 & 0.50 & -6.01 & \bf 24.35 & \bf 25.19\\

\hline
\end{tabular}
}
}
\end{center}

\label{tab:compare_input}
\vspace{-1em}
\end{table}

\smallvspace
\noindent \textbf{Strong Correlation with Sample BLEU.}\ \  
Our objective is strongly correlated with BLEU when using Input Sampling,
where $\Linput$ and $\LNAT$ should be balanced to achieve the best performance.
For example, Vanilla NAT does not introduce extra inputs, leading to large $\LNAT$;
CMLM introduces too many tokens in $Z$, bringing a large distortion from the original input.
However, $\hatLMPLE$ is less correlated with BLEU of Default Decoding, which can be caused by the decoding strategy as discussed below.

\smallvspace
\noindent \textbf{Potentials for Decoding Strategies.}\ \  \revise{Previous work \cite{glat2021qian} showed that CMLM performs poorly with a full masked decoder input, but we find that it can be improved by utilizing the input predictor to generate a better proxy input $Z$ in inference.
Specifically, Input Sampling brings about 4 BLEU points improvement on CMLM.} \rrevise{This idea is connected with the iterative NATs, where their refined sentence can be interpreted as a proxy input to improve the generation quality.}

\begin{figure}[!t]
  \centering
  \includegraphics[width=0.8\linewidth]{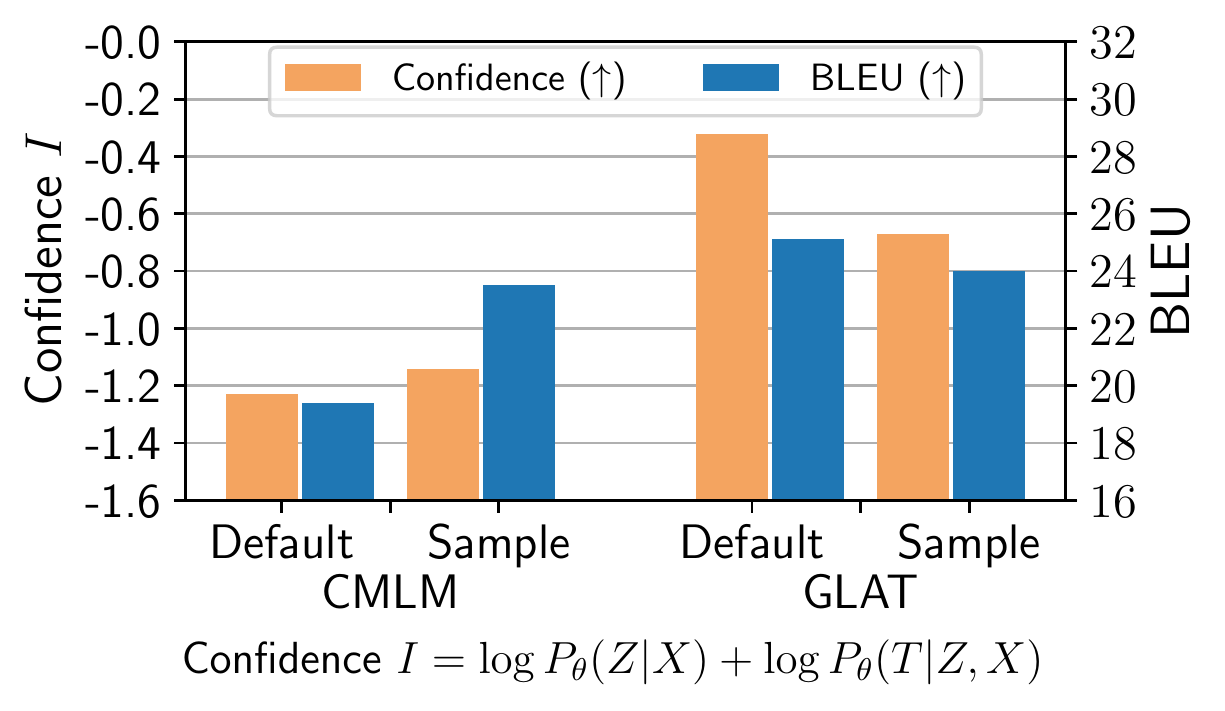}
  \vspace{-0.8em}
  \caption{\revise{Decoding Confidence $I$ and the BLEU score with two decoding strategies. CMLM with Input Sampling is more confident and thus achieves better BLEU than Default Decoding. GLAT is the opposite.}}
  \label{fig:decoding_confidence}
  \vspace{-1.5em}
\end{figure}

We also find that CMLM and GLAT prefer different decoding strategies, which can be explained by the decoding confidence $P_{\theta}(Z|X)$ and $P_{\theta}(T|Z, X)$. As shown in Fig.\ref{fig:decoding_confidence}, CMLM is more confident with Input Sampling than Default Decoding, whereas GLAT is the opposite.

\begin{table}[!b]
\vspace{-1em}
\caption{Comparing Dynamic KD against AT and previous NATs. \dag: Reported by \citet{glat2021qian} and \citet{oaxe2021du}. LPD \cite{imitatenat2019wei} and NPD \cite{nat2018gu} indicate reranking methods with the number of candidates. \revise{NPD are slower than LPD due to the use of an external AT reranker. Please see Table \ref{tab:compare_iterative} for more results.} }
\vspace{-0.8em}
\begin{center}
\resizebox{0.75\linewidth}{!}{
\setlength{\tabcolsep}{1mm}{

\begin{tabular}{c|l|ccc}
\hline
& \bf Models & \bf En-De & \bf Zh-En & \bf Speedup\\
\hline
AT & Transformer & 27.11 & 23.89 & 1.0x\\
\hline
\multirow{3}{*}{NAT} & MLE & 11.79 & 8.69 & 15.3x \\
& GLAT (NPD=7)\dag & 26.55 & / & 7.9x \\
& OaXE (LPD=5)\dag & 26.1 & 22.1 & 14.2x \\
\hline
\multirow{3}{*}{Ours} & GLAT + KD & 25.12 & 22.51 & 15.3x \\
& + Dynamic KD & 25.88 & 23.07 & 15.3x \\
& \ \ \ + LPD=3 & \bf 26.89 & \bf 24.42 & 14.6x \\
\hline
\end{tabular}
}
}
\end{center}
\label{tab:dynamic_kd}
\end{table}

\subsection{Results of Dynamic KD}
\label{sec:result-dynamic-kd}

\revise{We combine Dynamic KD with GLAT and further utilize a reranking method following our baselines.}
\revise{As shown in Table \ref{tab:dynamic_kd}, Dynamic KD brings about 0.6 $\sim$ 0.7 BLEU improvement against the single KD distilled from Transformer-\textit{base}.
Moreover, our best results achieve competitive translation quality with ATs with the modest cost in reranking.}

\begin{figure}[!t]
  \centering
  \includegraphics[width=0.9\linewidth]{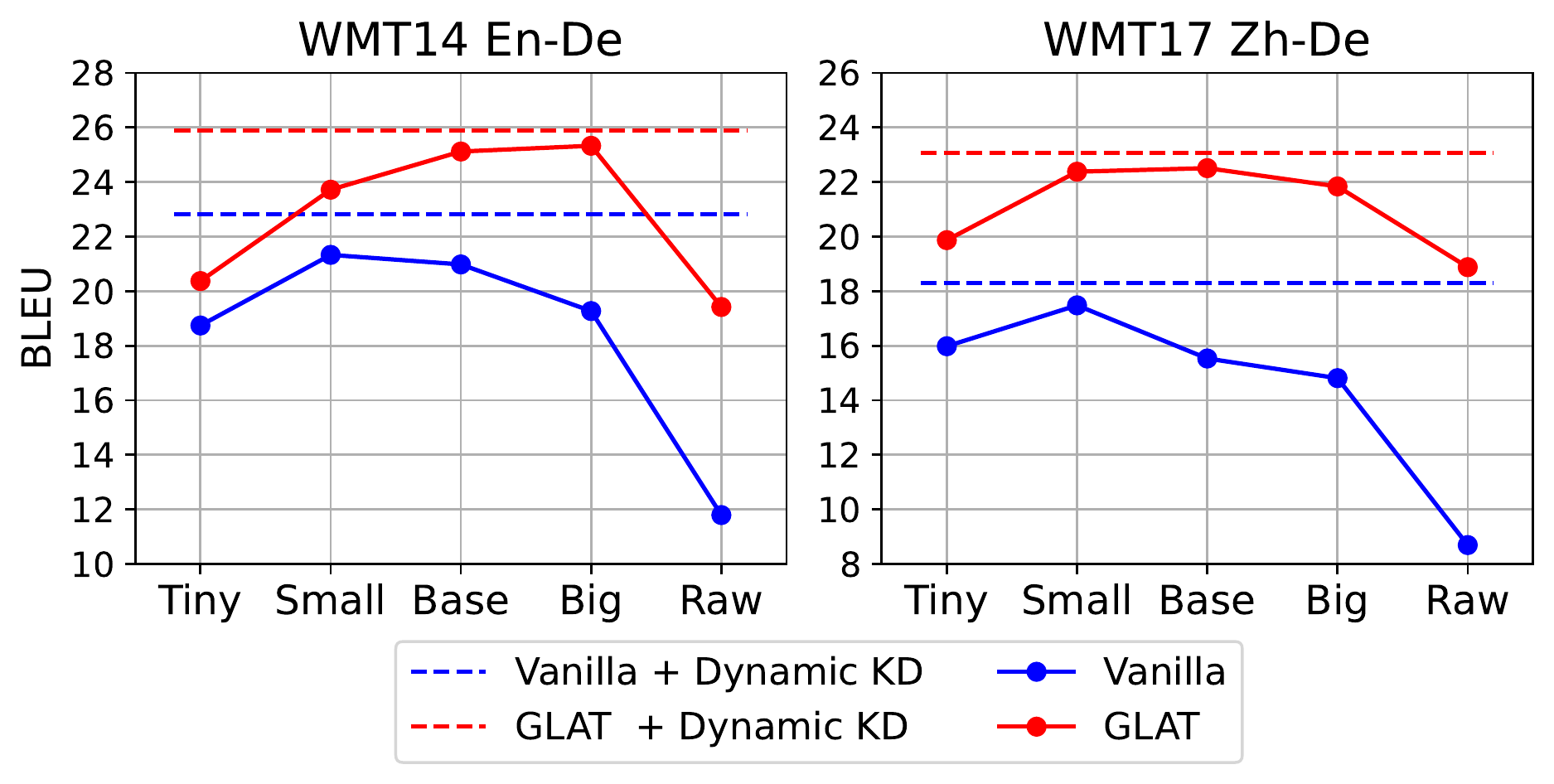}
  \vspace{-0.7em}
  \caption{Comparing Dynamic KD against NATs trained on single KD data. Tiny/Small/Base/Big indicates the size of \textit{AT teacher}. Raw represents the raw data. \revise{Dynamic KD utilizes all KD data and outperforms any single of them.}}
  \label{fig:dynamic_kd}
  \vspace{-1.5em}
\end{figure}

\revise{We further compare Dynamic KD against single KD data distilled from different AT teachers.
As shown in Fig.\ref{fig:dynamic_kd}, Dynamic KD outperforms the best result on any single KD data, verifying that explicitly balancing $\LNAT$ and the data distortion leads to better performance.
Notably, applying both proxy inputs and targets (GLAT+Dynamic KD) is better than simply applying one of them (Vanilla+Dynamic KD), showing that unifying the two methods of constructing proxy distributions is effective.}

The results suggest that our perspective effectively guides the design of new training methods. Explicitly optimizing $\LMPLE$ provides a promising way to find better proxy distributions, which outperforms existing heuristic methods.

\section{Related Work}

\rrevise{NATs are proposed to reduce the decoding latency but suffer from poor generation quality. 
Many studies are devoted to solving the problem.
Besides the methods discussed in our analysis, some studies are also helpful in improving the NAT performance, mainly including (1) objectives not based on cross-entropy \cite{imitatenat2019wei, crf2019sun, bagofngram2020shao, seqtrain2021shao}; (2) iteratively refining the generated outputs \cite{iterativerefinement2018lee, cmlm2019ghazvininejad, levenshtein2019gu, disco2020kasai, jmnat2020guo}.
Although the iterative approaches usually lead to better quality, \citet{deepshallow2021kasai} find that the these models are much slower and may not have advantages against ATs.
Moreover, recent works show that the non-iterative methods can also achieve competitive quality with AT models and have substantial lower latency than iterative methods~\cite{tricktrade2021gu, glatwmt2021qian, huang2022DATransformer}.}

\revise{Notably, a previous study \cite{kdnat2020zhou} also analyzes the NAT learning but mainly focuses on the KD method. They propose metrics to evaluate the complexity of the KD data and explain how KD improves NAT generation. Unlike their analysis that only considers the KD data, our perspective is more general in understanding many SoTA methods and better supported by the information theory.}\footnote{\revise{Specifically, their proposed metric may not correctly reflect the difficulties of NAT learning in some cases. For a dataset $y_1, \cdots, y_M \stackrel{iid}{\sim} p(\cdot|X)$, their proposed metric is high if $p(\cdot|X)$ has a large entropy. However, our $\tc=0$, correctly showing that the data satisfy the independent assumption.}}

\section{Conclusion}

\rrevise{
In this paper, we investigate the challenges in NAT learning.
From intuitive and theoretical perspectives, we show that the problem roots in the large information loss in capturing dependencies between tokens, where the dropped information can be measured by the dataset's \textit{conditional total correlation} $\tc$.
Furthermore, we revisit the existing successes in NAT learning and find that many previous studies alleviate the problem by maximizing the likelihood on a proxy distribution, which is designed to have a lower $\tc$.
Based on the analysis, we propose a unified framework named Maximum Proxy-Likelihood Estimation (MPLE), which provides a unified objective revealing how the choice of proxy distribution contributes to the final performance.
This framework improves our understanding of a wide range of NAT learning methods, including the SOTA ones like alignment-based objectives and glancing training.
Empirical analyses show that our perspective can well explain the phenomena in NAT learning, where the proposed objective highly correlates with the generation performance and can further guide the design of better training methods.
}

\section*{Acknowledgement}

\rrrevise{We would like to thank Yuxuan Song for the help in the derivation of Theorem 1 and the MPLE framework.} This work was supported by the National Science Foundation for Distinguished Young Scholars (with No. 62125604) and the NSFC projects (Key project with No. 61936010 and regular project with No. 61876096). This work was also supported by the Guoqiang Institute of Tsinghua University, with Grant No. 2019GQG1 and 2020GQG0005, and sponsored by Tsinghua-Toyota Joint Research Fund.

\nocite{tricktrade2021gu}
\nocite{levenshtein2019gu}
\nocite{emnatsun2020}
\nocite{CNAT2021bao}
\nocite{fairseq2019ott}
\nocite{dslp2021huang}

\bibliography{custom}

\begin{thebibliography}{45}
\providecommand{\natexlab}[1]{#1}
\providecommand{\url}[1]{\texttt{#1}}
\expandafter\ifx\csname urlstyle\endcsname\relax
  \providecommand{\doi}[1]{doi: #1}\else
  \providecommand{\doi}{doi: \begingroup \urlstyle{rm}\Url}\fi

\bibitem[Akaike(1998)]{mlekl1998akaike}
Akaike, H.
\newblock Information theory and an extension of the maximum likelihood
  principle.
\newblock In \emph{Selected papers of hirotugu akaike}, pp.\  199--213.
  Springer, 1998.

\bibitem[Bao et~al.(2019)Bao, Zhou, Feng, Wang, Huang, Chen, and
  Li]{pnat2019bao}
Bao, Y., Zhou, H., Feng, J., Wang, M., Huang, S., Chen, J., and Li, L.
\newblock Non-autoregressive transformer by position learning.
\newblock \emph{CoRR}, abs/1911.10677, 2019.
\newblock URL \url{http://arxiv.org/abs/1911.10677}.

\bibitem[Bao et~al.(2021)Bao, Huang, Xiao, Wang, Dai, and Chen]{CNAT2021bao}
Bao, Y., Huang, S., Xiao, T., Wang, D., Dai, X., and Chen, J.
\newblock Non-autoregressive translation by learning target categorical codes.
\newblock In Toutanova, K., Rumshisky, A., Zettlemoyer, L.,
  Hakkani{-}T{\"{u}}r, D., Beltagy, I., Bethard, S., Cotterell, R.,
  Chakraborty, T., and Zhou, Y. (eds.), \emph{Proceedings of the 2021
  Conference of the North American Chapter of the Association for Computational
  Linguistics: Human Language Technologies, {NAACL-HLT} 2021, Online, June
  6-11, 2021}, pp.\  5749--5759. Association for Computational Linguistics,
  2021.
\newblock \doi{10.18653/v1/2021.naacl-main.458}.
\newblock URL \url{https://doi.org/10.18653/v1/2021.naacl-main.458}.

\bibitem[Bao et~al.(2022)Bao, Zhou, Huang, Wang, Qian, Dai, Chen, and
  Li]{latentglat2022}
Bao, Y., Zhou, H., Huang, S., Wang, D., Qian, L., Dai, X., Chen, J., and Li, L.
\newblock latent-glat: Glancing at latent variables for parallel text
  generation.
\newblock \emph{CoRR}, abs/2204.02030, 2022.
\newblock \doi{10.48550/arXiv.2204.02030}.
\newblock URL \url{https://doi.org/10.48550/arXiv.2204.02030}.

\bibitem[Ding et~al.(2021{\natexlab{a}})Ding, Wang, Liu, Wong, Tao, and
  Tu]{lexical2021ding}
Ding, L., Wang, L., Liu, X., Wong, D.~F., Tao, D., and Tu, Z.
\newblock Understanding and improving lexical choice in non-autoregressive
  translation.
\newblock In \emph{9th International Conference on Learning Representations,
  {ICLR} 2021, Virtual Event, Austria, May 3-7, 2021}. OpenReview.net,
  2021{\natexlab{a}}.
\newblock URL \url{https://openreview.net/forum?id=ZTFeSBIX9C}.

\bibitem[Ding et~al.(2021{\natexlab{b}})Ding, Wang, Liu, Wong, Tao, and
  Tu]{reversekd2021ding}
Ding, L., Wang, L., Liu, X., Wong, D.~F., Tao, D., and Tu, Z.
\newblock Rejuvenating low-frequency words: Making the most of parallel data in
  non-autoregressive translation.
\newblock In Zong, C., Xia, F., Li, W., and Navigli, R. (eds.),
  \emph{Proceedings of the 59th Annual Meeting of the Association for
  Computational Linguistics and the 11th International Joint Conference on
  Natural Language Processing, {ACL/IJCNLP} 2021, (Volume 1: Long Papers),
  Virtual Event, August 1-6, 2021}, pp.\  3431--3441. Association for
  Computational Linguistics, 2021{\natexlab{b}}.
\newblock \doi{10.18653/v1/2021.acl-long.266}.
\newblock URL \url{https://doi.org/10.18653/v1/2021.acl-long.266}.

\bibitem[Du et~al.(2021)Du, Tu, and Jiang]{oaxe2021du}
Du, C., Tu, Z., and Jiang, J.
\newblock Order-agnostic cross entropy for non-autoregressive machine
  translation.
\newblock In Meila, M. and Zhang, T. (eds.), \emph{Proceedings of the 38th
  International Conference on Machine Learning, {ICML} 2021, 18-24 July 2021,
  Virtual Event}, volume 139 of \emph{Proceedings of Machine Learning
  Research}, pp.\  2849--2859. {PMLR}, 2021.
\newblock URL \url{http://proceedings.mlr.press/v139/du21c.html}.

\bibitem[Fox \& Roberts(2012)Fox and Roberts]{variaionaltutorial2012fox}
Fox, C.~W. and Roberts, S.~J.
\newblock A tutorial on variational bayesian inference.
\newblock \emph{Artif. Intell. Rev.}, 38\penalty0 (2):\penalty0 85--95, 2012.
\newblock \doi{10.1007/s10462-011-9236-8}.
\newblock URL \url{https://doi.org/10.1007/s10462-011-9236-8}.

\bibitem[Ghazvininejad et~al.(2019)Ghazvininejad, Levy, Liu, and
  Zettlemoyer]{cmlm2019ghazvininejad}
Ghazvininejad, M., Levy, O., Liu, Y., and Zettlemoyer, L.
\newblock Mask-predict: Parallel decoding of conditional masked language
  models.
\newblock In Inui, K., Jiang, J., Ng, V., and Wan, X. (eds.), \emph{Proceedings
  of the 2019 Conference on Empirical Methods in Natural Language Processing
  and the 9th International Joint Conference on Natural Language Processing,
  {EMNLP-IJCNLP} 2019, Hong Kong, China, November 3-7, 2019}, pp.\  6111--6120.
  Association for Computational Linguistics, 2019.
\newblock \doi{10.18653/v1/D19-1633}.
\newblock URL \url{https://doi.org/10.18653/v1/D19-1633}.

\bibitem[Ghazvininejad et~al.(2020)Ghazvininejad, Karpukhin, Zettlemoyer, and
  Levy]{axe2020ghazvininejad}
Ghazvininejad, M., Karpukhin, V., Zettlemoyer, L., and Levy, O.
\newblock Aligned cross entropy for non-autoregressive machine translation.
\newblock In \emph{Proceedings of the 37th International Conference on Machine
  Learning, {ICML} 2020, 13-18 July 2020, Virtual Event}, volume 119 of
  \emph{Proceedings of Machine Learning Research}, pp.\  3515--3523. {PMLR},
  2020.
\newblock URL \url{http://proceedings.mlr.press/v119/ghazvininejad20a.html}.

\bibitem[Gu \& Kong(2021)Gu and Kong]{tricktrade2021gu}
Gu, J. and Kong, X.
\newblock Fully non-autoregressive neural machine translation: Tricks of the
  trade.
\newblock In Zong, C., Xia, F., Li, W., and Navigli, R. (eds.), \emph{Findings
  of the Association for Computational Linguistics: {ACL/IJCNLP} 2021, Online
  Event, August 1-6, 2021}, volume {ACL/IJCNLP} 2021 of \emph{Findings of
  {ACL}}, pp.\  120--133. Association for Computational Linguistics, 2021.
\newblock \doi{10.18653/v1/2021.findings-acl.11}.
\newblock URL \url{https://doi.org/10.18653/v1/2021.findings-acl.11}.

\bibitem[Gu et~al.(2018)Gu, Bradbury, Xiong, Li, and Socher]{nat2018gu}
Gu, J., Bradbury, J., Xiong, C., Li, V. O.~K., and Socher, R.
\newblock Non-autoregressive neural machine translation.
\newblock In \emph{6th International Conference on Learning Representations,
  {ICLR} 2018, Vancouver, BC, Canada, April 30 - May 3, 2018, Conference Track
  Proceedings}. OpenReview.net, 2018.
\newblock URL \url{https://openreview.net/forum?id=B1l8BtlCb}.

\bibitem[Gu et~al.(2019)Gu, Wang, and Zhao]{levenshtein2019gu}
Gu, J., Wang, C., and Zhao, J.
\newblock Levenshtein transformer.
\newblock In Wallach, H.~M., Larochelle, H., Beygelzimer, A.,
  d'Alch{\'{e}}{-}Buc, F., Fox, E.~B., and Garnett, R. (eds.), \emph{Advances
  in Neural Information Processing Systems 32: Annual Conference on Neural
  Information Processing Systems 2019, NeurIPS 2019, December 8-14, 2019,
  Vancouver, BC, Canada}, pp.\  11179--11189, 2019.
\newblock URL
  \url{https://proceedings.neurips.cc/paper/2019/hash/675f9820626f5bc0afb47b57890b466e-Abstract.html}.

\bibitem[Guo et~al.(2020)Guo, Xu, and Chen]{jmnat2020guo}
Guo, J., Xu, L., and Chen, E.
\newblock Jointly masked sequence-to-sequence model for non-autoregressive
  neural machine translation.
\newblock In Jurafsky, D., Chai, J., Schluter, N., and Tetreault, J.~R. (eds.),
  \emph{Proceedings of the 58th Annual Meeting of the Association for
  Computational Linguistics, {ACL} 2020, Online, July 5-10, 2020}, pp.\
  376--385. Association for Computational Linguistics, 2020.
\newblock \doi{10.18653/v1/2020.acl-main.36}.
\newblock URL \url{https://doi.org/10.18653/v1/2020.acl-main.36}.

\bibitem[Hassan et~al.(2018)Hassan, Aue, Chen, Chowdhary, Clark, Federmann,
  Huang, Junczys{-}Dowmunt, Lewis, Li, Liu, Liu, Luo, Menezes, Qin, Seide, Tan,
  Tian, Wu, Wu, Xia, Zhang, Zhang, and Zhou]{parity2018hassan}
Hassan, H., Aue, A., Chen, C., Chowdhary, V., Clark, J., Federmann, C., Huang,
  X., Junczys{-}Dowmunt, M., Lewis, W., Li, M., Liu, S., Liu, T., Luo, R.,
  Menezes, A., Qin, T., Seide, F., Tan, X., Tian, F., Wu, L., Wu, S., Xia, Y.,
  Zhang, D., Zhang, Z., and Zhou, M.
\newblock Achieving human parity on automatic chinese to english news
  translation.
\newblock \emph{CoRR}, abs/1803.05567, 2018.
\newblock URL \url{http://arxiv.org/abs/1803.05567}.

\bibitem[Huang et~al.(2022{\natexlab{a}})Huang, Zhou, Za{\"{\i}}ane, Mou, and
  Li]{dslp2021huang}
Huang, C., Zhou, H., Za{\"{\i}}ane, O.~R., Mou, L., and Li, L.
\newblock Non-autoregressive translation with layer-wise prediction and deep
  supervision.
\newblock \emph{The Thirty-Sixth {AAAI} Conference on Artificial Intelligence,
  {AAAI} 2022}, 2022{\natexlab{a}}.
\newblock URL \url{https://arxiv.org/abs/2110.07515}.

\bibitem[Huang et~al.(2022{\natexlab{b}})Huang, Zhou, Liu, Li, and
  Huang]{huang2022DATransformer}
Huang, F., Zhou, H., Liu, Y., Li, H., and Huang, M.
\newblock Directed acyclic transformer for non-autoregressive machine
  translation.
\newblock In \emph{Proceedings of the 39th International Conference on Machine
  Learning, {ICML} 2022}, 2022{\natexlab{b}}.
\newblock URL \url{https://arxiv.org/abs/2205.07459}.

\bibitem[Kaiser et~al.(2018)Kaiser, Bengio, Roy, Vaswani, Parmar, Uszkoreit,
  and Shazeer]{discretelatent2018kaiser}
Kaiser, L., Bengio, S., Roy, A., Vaswani, A., Parmar, N., Uszkoreit, J., and
  Shazeer, N.
\newblock Fast decoding in sequence models using discrete latent variables.
\newblock In Dy, J.~G. and Krause, A. (eds.), \emph{Proceedings of the 35th
  International Conference on Machine Learning, {ICML} 2018,
  Stockholmsm{\"{a}}ssan, Stockholm, Sweden, July 10-15, 2018}, volume~80 of
  \emph{Proceedings of Machine Learning Research}, pp.\  2395--2404. {PMLR},
  2018.
\newblock URL \url{http://proceedings.mlr.press/v80/kaiser18a.html}.

\bibitem[Kasai et~al.(2020)Kasai, Cross, Ghazvininejad, and Gu]{disco2020kasai}
Kasai, J., Cross, J., Ghazvininejad, M., and Gu, J.
\newblock Non-autoregressive machine translation with disentangled context
  transformer.
\newblock In \emph{Proceedings of the 37th International Conference on Machine
  Learning, {ICML} 2020, 13-18 July 2020, Virtual Event}, volume 119 of
  \emph{Proceedings of Machine Learning Research}, pp.\  5144--5155. {PMLR},
  2020.
\newblock URL \url{http://proceedings.mlr.press/v119/kasai20a.html}.

\bibitem[Kasai et~al.(2021)Kasai, Pappas, Peng, Cross, and
  Smith]{deepshallow2021kasai}
Kasai, J., Pappas, N., Peng, H., Cross, J., and Smith, N.~A.
\newblock Deep encoder, shallow decoder: Reevaluating non-autoregressive
  machine translation.
\newblock In \emph{9th International Conference on Learning Representations,
  {ICLR} 2021, Virtual Event, Austria, May 3-7, 2021}. OpenReview.net, 2021.
\newblock URL \url{https://openreview.net/forum?id=KpfasTaLUpq}.

\bibitem[Kim \& Rush(2016)Kim and Rush]{seqkd2016kim}
Kim, Y. and Rush, A.~M.
\newblock Sequence-level knowledge distillation.
\newblock In Su, J., Carreras, X., and Duh, K. (eds.), \emph{Proceedings of the
  2016 Conference on Empirical Methods in Natural Language Processing, {EMNLP}
  2016, Austin, Texas, USA, November 1-4, 2016}, pp.\  1317--1327. The
  Association for Computational Linguistics, 2016.
\newblock \doi{10.18653/v1/d16-1139}.
\newblock URL \url{https://doi.org/10.18653/v1/d16-1139}.

\bibitem[Lee et~al.(2018)Lee, Mansimov, and Cho]{iterativerefinement2018lee}
Lee, J., Mansimov, E., and Cho, K.
\newblock Deterministic non-autoregressive neural sequence modeling by
  iterative refinement.
\newblock In Riloff, E., Chiang, D., Hockenmaier, J., and Tsujii, J. (eds.),
  \emph{Proceedings of the 2018 Conference on Empirical Methods in Natural
  Language Processing, Brussels, Belgium, October 31 - November 4, 2018}, pp.\
  1173--1182. Association for Computational Linguistics, 2018.
\newblock \doi{10.18653/v1/d18-1149}.
\newblock URL \url{https://doi.org/10.18653/v1/d18-1149}.

\bibitem[Libovick{\'{y}} \& Helcl(2018)Libovick{\'{y}} and
  Helcl]{ctc2018libovicky}
Libovick{\'{y}}, J. and Helcl, J.
\newblock End-to-end non-autoregressive neural machine translation with
  connectionist temporal classification.
\newblock In Riloff, E., Chiang, D., Hockenmaier, J., and Tsujii, J. (eds.),
  \emph{Proceedings of the 2018 Conference on Empirical Methods in Natural
  Language Processing, Brussels, Belgium, October 31 - November 4, 2018}, pp.\
  3016--3021. Association for Computational Linguistics, 2018.
\newblock \doi{10.18653/v1/d18-1336}.
\newblock URL \url{https://doi.org/10.18653/v1/d18-1336}.

\bibitem[Ma et~al.(2019)Ma, Zhou, Li, Neubig, and Hovy]{flowseq2019ma}
Ma, X., Zhou, C., Li, X., Neubig, G., and Hovy, E.~H.
\newblock Flowseq: Non-autoregressive conditional sequence generation with
  generative flow.
\newblock In Inui, K., Jiang, J., Ng, V., and Wan, X. (eds.), \emph{Proceedings
  of the 2019 Conference on Empirical Methods in Natural Language Processing
  and the 9th International Joint Conference on Natural Language Processing,
  {EMNLP-IJCNLP} 2019, Hong Kong, China, November 3-7, 2019}, pp.\  4281--4291.
  Association for Computational Linguistics, 2019.
\newblock \doi{10.18653/v1/D19-1437}.
\newblock URL \url{https://doi.org/10.18653/v1/D19-1437}.

\bibitem[Ott et~al.(2018)Ott, Auli, Grangier, and Ranzato]{uncertainty2018ott}
Ott, M., Auli, M., Grangier, D., and Ranzato, M.
\newblock Analyzing uncertainty in neural machine translation.
\newblock In Dy, J.~G. and Krause, A. (eds.), \emph{Proceedings of the 35th
  International Conference on Machine Learning, {ICML} 2018,
  Stockholmsm{\"{a}}ssan, Stockholm, Sweden, July 10-15, 2018}, volume~80 of
  \emph{Proceedings of Machine Learning Research}, pp.\  3953--3962. {PMLR},
  2018.
\newblock URL \url{http://proceedings.mlr.press/v80/ott18a.html}.

\bibitem[Ott et~al.(2019)Ott, Edunov, Baevski, Fan, Gross, Ng, Grangier, and
  Auli]{fairseq2019ott}
Ott, M., Edunov, S., Baevski, A., Fan, A., Gross, S., Ng, N., Grangier, D., and
  Auli, M.
\newblock fairseq: {A} fast, extensible toolkit for sequence modeling.
\newblock In Ammar, W., Louis, A., and Mostafazadeh, N. (eds.),
  \emph{Proceedings of the 2019 Conference of the North American Chapter of the
  Association for Computational Linguistics: Human Language Technologies,
  {NAACL-HLT} 2019, Minneapolis, MN, USA, June 2-7, 2019, Demonstrations}, pp.\
   48--53. Association for Computational Linguistics, 2019.
\newblock \doi{10.18653/v1/n19-4009}.
\newblock URL \url{https://doi.org/10.18653/v1/n19-4009}.

\bibitem[Papineni et~al.(2002)Papineni, Roukos, Ward, and
  Zhu]{bleu2002papineni}
Papineni, K., Roukos, S., Ward, T., and Zhu, W.
\newblock Bleu: a method for automatic evaluation of machine translation.
\newblock In \emph{Proceedings of the 40th Annual Meeting of the Association
  for Computational Linguistics, July 6-12, 2002, Philadelphia, PA, {USA}},
  pp.\  311--318. {ACL}, 2002.
\newblock \doi{10.3115/1073083.1073135}.
\newblock URL \url{https://aclanthology.org/P02-1040/}.

\bibitem[Qian et~al.(2021{\natexlab{a}})Qian, Zhou, Bao, Wang, Qiu, Zhang, Yu,
  and Li]{glat2021qian}
Qian, L., Zhou, H., Bao, Y., Wang, M., Qiu, L., Zhang, W., Yu, Y., and Li, L.
\newblock Glancing transformer for non-autoregressive neural machine
  translation.
\newblock In Zong, C., Xia, F., Li, W., and Navigli, R. (eds.),
  \emph{Proceedings of the 59th Annual Meeting of the Association for
  Computational Linguistics and the 11th International Joint Conference on
  Natural Language Processing, {ACL/IJCNLP} 2021, (Volume 1: Long Papers),
  Virtual Event, August 1-6, 2021}, pp.\  1993--2003. Association for
  Computational Linguistics, 2021{\natexlab{a}}.
\newblock URL \url{https://aclanthology.org/2021.acl-long.155}.

\bibitem[Qian et~al.(2021{\natexlab{b}})Qian, Zhou, Zheng, Zhu, Lin, Feng,
  Cheng, Li, Wang, and Zhou]{glatwmt2021qian}
Qian, L., Zhou, Y., Zheng, Z., Zhu, Y., Lin, Z., Feng, J., Cheng, S., Li, L.,
  Wang, M., and Zhou, H.
\newblock The volctrans {GLAT} system: Non-autoregressive translation meets
  {WMT21}.
\newblock \emph{CoRR}, abs/2109.11247, 2021{\natexlab{b}}.
\newblock URL \url{https://arxiv.org/abs/2109.11247}.

\bibitem[Ran et~al.(2021)Ran, Lin, Li, and Zhou]{reorder2021ran}
Ran, Q., Lin, Y., Li, P., and Zhou, J.
\newblock Guiding non-autoregressive neural machine translation decoding with
  reordering information.
\newblock In \emph{Thirty-Fifth {AAAI} Conference on Artificial Intelligence,
  {AAAI} 2021, Thirty-Third Conference on Innovative Applications of Artificial
  Intelligence, {IAAI} 2021, The Eleventh Symposium on Educational Advances in
  Artificial Intelligence, {EAAI} 2021, Virtual Event, February 2-9, 2021},
  pp.\  13727--13735. {AAAI} Press, 2021.
\newblock URL \url{https://ojs.aaai.org/index.php/AAAI/article/view/17618}.

\bibitem[Saharia et~al.(2020)Saharia, Chan, Saxena, and
  Norouzi]{imputermt2020saharia}
Saharia, C., Chan, W., Saxena, S., and Norouzi, M.
\newblock Non-autoregressive machine translation with latent alignments.
\newblock In Webber, B., Cohn, T., He, Y., and Liu, Y. (eds.),
  \emph{Proceedings of the 2020 Conference on Empirical Methods in Natural
  Language Processing, {EMNLP} 2020, Online, November 16-20, 2020}, pp.\
  1098--1108. Association for Computational Linguistics, 2020.
\newblock \doi{10.18653/v1/2020.emnlp-main.83}.
\newblock URL \url{https://doi.org/10.18653/v1/2020.emnlp-main.83}.

\bibitem[Sennrich et~al.(2016)Sennrich, Haddow, and Birch]{bpe2016sennrich}
Sennrich, R., Haddow, B., and Birch, A.
\newblock Neural machine translation of rare words with subword units.
\newblock In \emph{Proceedings of the 54th Annual Meeting of the Association
  for Computational Linguistics, {ACL} 2016, August 7-12, 2016, Berlin,
  Germany, Volume 1: Long Papers}. The Association for Computer Linguistics,
  2016.
\newblock \doi{10.18653/v1/p16-1162}.
\newblock URL \url{https://doi.org/10.18653/v1/p16-1162}.

\bibitem[Shao et~al.(2020)Shao, Zhang, Feng, Meng, and
  Zhou]{bagofngram2020shao}
Shao, C., Zhang, J., Feng, Y., Meng, F., and Zhou, J.
\newblock Minimizing the bag-of-ngrams difference for non-autoregressive neural
  machine translation.
\newblock In \emph{The Thirty-Fourth {AAAI} Conference on Artificial
  Intelligence, {AAAI} 2020, The Thirty-Second Innovative Applications of
  Artificial Intelligence Conference, {IAAI} 2020, The Tenth {AAAI} Symposium
  on Educational Advances in Artificial Intelligence, {EAAI} 2020, New York,
  NY, USA, February 7-12, 2020}, pp.\  198--205. {AAAI} Press, 2020.
\newblock URL \url{https://aaai.org/ojs/index.php/AAAI/article/view/5351}.

\bibitem[Shao et~al.(2021)Shao, Feng, Zhang, Meng, and Zhou]{seqtrain2021shao}
Shao, C., Feng, Y., Zhang, J., Meng, F., and Zhou, J.
\newblock Sequence-level training for non-autoregressive neural machine
  translation.
\newblock \emph{Comput. Linguistics}, 47\penalty0 (4):\penalty0 891--925, 2021.
\newblock \doi{10.1162/coli\_a\_00421}.
\newblock URL \url{https://doi.org/10.1162/coli\_a\_00421}.

\bibitem[Shu et~al.(2020)Shu, Lee, Nakayama, and Cho]{vae2020shu}
Shu, R., Lee, J., Nakayama, H., and Cho, K.
\newblock Latent-variable non-autoregressive neural machine translation with
  deterministic inference using a delta posterior.
\newblock In \emph{The Thirty-Fourth {AAAI} Conference on Artificial
  Intelligence, {AAAI} 2020, The Thirty-Second Innovative Applications of
  Artificial Intelligence Conference, {IAAI} 2020, The Tenth {AAAI} Symposium
  on Educational Advances in Artificial Intelligence, {EAAI} 2020, New York,
  NY, USA, February 7-12, 2020}, pp.\  8846--8853. {AAAI} Press, 2020.
\newblock URL \url{https://aaai.org/ojs/index.php/AAAI/article/view/6413}.

\bibitem[Studen{\'{y}} \& Vejnarov{\'{a}}(1998)Studen{\'{y}} and
  Vejnarov{\'{a}}]{multiinfomration1998studeny}
Studen{\'{y}}, M. and Vejnarov{\'{a}}, J.
\newblock The multiinformation function as a tool for measuring stochastic
  dependence.
\newblock In Jordan, M.~I. (ed.), \emph{Learning in Graphical Models},
  volume~89 of \emph{{NATO} {ASI} Series}, pp.\  261--297. Springer
  Netherlands, 1998.
\newblock \doi{10.1007/978-94-011-5014-9\_10}.
\newblock URL \url{https://doi.org/10.1007/978-94-011-5014-9\_10}.

\bibitem[Sun \& Yang(2020)Sun and Yang]{emnatsun2020}
Sun, Z. and Yang, Y.
\newblock An {EM} approach to non-autoregressive conditional sequence
  generation.
\newblock In \emph{Proceedings of the 37th International Conference on Machine
  Learning, {ICML} 2020, 13-18 July 2020, Virtual Event}, volume 119 of
  \emph{Proceedings of Machine Learning Research}, pp.\  9249--9258. {PMLR},
  2020.
\newblock URL \url{http://proceedings.mlr.press/v119/sun20c.html}.

\bibitem[Sun et~al.(2019)Sun, Li, Wang, He, Lin, and Deng]{crf2019sun}
Sun, Z., Li, Z., Wang, H., He, D., Lin, Z., and Deng, Z.
\newblock Fast structured decoding for sequence models.
\newblock In Wallach, H.~M., Larochelle, H., Beygelzimer, A.,
  d'Alch{\'{e}}{-}Buc, F., Fox, E.~B., and Garnett, R. (eds.), \emph{Advances
  in Neural Information Processing Systems 32: Annual Conference on Neural
  Information Processing Systems 2019, NeurIPS 2019, December 8-14, 2019,
  Vancouver, BC, Canada}, pp.\  3011--3020, 2019.
\newblock URL
  \url{https://proceedings.neurips.cc/paper/2019/hash/74563ba21a90da13dacf2a73e3ddefa7-Abstract.html}.

\bibitem[Vaswani et~al.(2017)Vaswani, Shazeer, Parmar, Uszkoreit, Jones, Gomez,
  Kaiser, and Polosukhin]{transformer2017vaswani}
Vaswani, A., Shazeer, N., Parmar, N., Uszkoreit, J., Jones, L., Gomez, A.~N.,
  Kaiser, L., and Polosukhin, I.
\newblock Attention is all you need.
\newblock In Guyon, I., von Luxburg, U., Bengio, S., Wallach, H.~M., Fergus,
  R., Vishwanathan, S. V.~N., and Garnett, R. (eds.), \emph{Advances in Neural
  Information Processing Systems 30: Annual Conference on Neural Information
  Processing Systems 2017, December 4-9, 2017, Long Beach, CA, {USA}}, pp.\
  5998--6008, 2017.
\newblock URL
  \url{https://proceedings.neurips.cc/paper/2017/hash/3f5ee243547dee91fbd053c1c4a845aa-Abstract.html}.

\bibitem[Watanabe(1960)]{totalcorrelation1960watanabe}
Watanabe, M.~S.
\newblock Information theoretical analysis of multivariate correlation.
\newblock \emph{{IBM} J. Res. Dev.}, 4\penalty0 (1):\penalty0 66--82, 1960.
\newblock \doi{10.1147/rd.41.0066}.
\newblock URL \url{https://doi.org/10.1147/rd.41.0066}.

\bibitem[Wei et~al.(2019{\natexlab{a}})Wei, Wang, Zhou, Lin, and
  Sun]{imitatenat2019wei}
Wei, B., Wang, M., Zhou, H., Lin, J., and Sun, X.
\newblock Imitation learning for non-autoregressive neural machine translation.
\newblock In Korhonen, A., Traum, D.~R., and M{\`{a}}rquez, L. (eds.),
  \emph{Proceedings of the 57th Conference of the Association for Computational
  Linguistics, {ACL} 2019, Florence, Italy, July 28- August 2, 2019, Volume 1:
  Long Papers}, pp.\  1304--1312. Association for Computational Linguistics,
  2019{\natexlab{a}}.
\newblock \doi{10.18653/v1/p19-1125}.
\newblock URL \url{https://doi.org/10.18653/v1/p19-1125}.

\bibitem[Wei et~al.(2019{\natexlab{b}})Wei, Wang, Zhou, Lin, and
  Sun]{imitation2019wei}
Wei, B., Wang, M., Zhou, H., Lin, J., and Sun, X.
\newblock Imitation learning for non-autoregressive neural machine translation.
\newblock In Korhonen, A., Traum, D.~R., and M{\`{a}}rquez, L. (eds.),
  \emph{Proceedings of the 57th Conference of the Association for Computational
  Linguistics, {ACL} 2019, Florence, Italy, July 28- August 2, 2019, Volume 1:
  Long Papers}, pp.\  1304--1312. Association for Computational Linguistics,
  2019{\natexlab{b}}.
\newblock \doi{10.18653/v1/p19-1125}.
\newblock URL \url{https://doi.org/10.18653/v1/p19-1125}.

\bibitem[Xu et~al.(2020)Xu, Zhao, Song, Stewart, and Ermon]{vinformation2020xu}
Xu, Y., Zhao, S., Song, J., Stewart, R., and Ermon, S.
\newblock A theory of usable information under computational constraints.
\newblock In \emph{8th International Conference on Learning Representations,
  {ICLR} 2020, Addis Ababa, Ethiopia, April 26-30, 2020}. OpenReview.net, 2020.
\newblock URL \url{https://openreview.net/forum?id=r1eBeyHFDH}.

\bibitem[Yang et~al.(2021)Yang, Lei, Liu, Qi, and Lv]{posconstrain2021yang}
Yang, K., Lei, W., Liu, D., Qi, W., and Lv, J.
\newblock Pos-constrained parallel decoding for non-autoregressive generation.
\newblock In Zong, C., Xia, F., Li, W., and Navigli, R. (eds.),
  \emph{Proceedings of the 59th Annual Meeting of the Association for
  Computational Linguistics and the 11th International Joint Conference on
  Natural Language Processing, {ACL/IJCNLP} 2021, (Volume 1: Long Papers),
  Virtual Event, August 1-6, 2021}, pp.\  5990--6000. Association for
  Computational Linguistics, 2021.
\newblock \doi{10.18653/v1/2021.acl-long.467}.
\newblock URL \url{https://doi.org/10.18653/v1/2021.acl-long.467}.

\bibitem[Zhou et~al.(2020)Zhou, Gu, and Neubig]{kdnat2020zhou}
Zhou, C., Gu, J., and Neubig, G.
\newblock Understanding knowledge distillation in non-autoregressive machine
  translation.
\newblock In \emph{8th International Conference on Learning Representations,
  {ICLR} 2020, Addis Ababa, Ethiopia, April 26-30, 2020}. OpenReview.net, 2020.
\newblock URL \url{https://openreview.net/forum?id=BygFVAEKDH}.

\end{thebibliography}
\bibliographystyle{icml2022}

\clearpage
\appendix
\onecolumn

\section{Relation to Iterative NATs}
\label{app:iterative}

Although MPLE provides a unified perspective to understand many previous methods, we do not discuss an important branch of the NAT model, i.e., the iterative NATs.
The reason lies in the conditional independent assumption of Eq.\ref{eq:nat_ll}, which is the basic assumption of our analyses but not satisfied in the iterative methods.
However, our perspective can also improve the understanding of iterative NATs, and we point out some important relations.

\textbf{$\tc$ measures the information loss in each iterative step.} Although iterative NATs do not satisfy Eq.\ref{eq:nat_ll}, they still predict tokens independently in each refinement step, so the minimum information loss can be measured by $\tc$.
For example, in some popular iterative NATs \cite{iterativerefinement2018lee, cmlm2019ghazvininejad}, the $i$-th refinement step's log-likelihood is defined as
\begin{align}
    \log P_{\theta}^{\text{NAT}}(Y|X, Y^{(i-1)}) = \sum_{k=1}^M \log P_{\theta}^{\text{NAT}}(y_k|X, Y^{(i-1)}),
\end{align}
where $Y$ is the target sentence, $Y^{(i-1)}$ is the refined result of previous steps, and $M$ is the target length. Similar to Theorem 1, we can prove that the minimal KL divergence is $\tc_i$, i.e., the conditional TC of the target distribution when $Y^{(i-1)}$ is given, where
\begin{align}
\tc_i = \sum_{k=1}^{M} H_{\text{data}}(y_k|X, Y^{(i-1)}) - H_{\text{data}}(Y|X, Y^{(i-1)}),
\end{align}
The result implies that iterative NATs also suffer from the information loss due to the dependency dropping and explains why they can benefit from methods that reduces $\tc$, e.g., knowledge distillation.

\textbf{Iterative Masked Prediction are special cases of MPLE with shared Input Predictor and NAT decoder.}
If we regard the output of the ($i-1$)-th refinement as the proxy input, iterative NATs actually construct a proxy distribution $\Pdata(Y|X, Y^{(i-1)})$, which reduces $\tc$ by providing an extra decoder input. Unlike the input predictor defined in our MPLE framework, iterative NATs predict the proxy input $Y^{(i-1)}$ by the NAT decoder itself with $i-1$ refinement steps. In Table.\ref{tab:compare_target}, we propose Input Sampling method that generates $Z$ from the input predictor, which is similar to a single step refinement but with a separate NAT decoder.

\section{Conditional TC and Performance Gap}

\label{app:cgap}

Table \ref{tab:total-correlation} aims to show that the large $\tc$ is the main obstacle in NAT learning, and we provide more details here.

\textbf{Dataset}\ \ \  We first choose WMT14 En-De and WMT16 En-Ro, which contains 4.5M pairs and 610k pairs in the training set, respectively. Since natural datasets usually have a large $\tc$, we further construct two synthetic datasets for comparison. Both synthetic datasets use the English corpus in WMT14 En-De as targets, and the source sentences are modified from the targets by word replacement or word dropping. In other words, the synthetic dataset trains the NAT to generate clean English sentences from corrupted English inputs.
Specifically, \textit{Synthetic A} replaces 50\% of tokens by randomly sampled tokens from the vocabulary. \textit{Synthetic B} further drops 10\% of tokens in the source sentences based on Synthetic A.

\textbf{Estimation of Conditional TC}\ \ \ To estimate $\tc$, we use \textit{V-entropy} \cite{vinformation2020xu} instead of the Shannon entropy because the latter is intractable due to the unknown data distribution. The V-entropy is comparable only when the function family for estimation is fixed. In our implementation, we use Transformers-\textit{base} as the function family.

\textbf{More Rigorous Comparison}\ \ \ Since BLEU is not strictly comparable across datasets, we present a more rigorous comparison by estimating the parameter size required for an autoregressive Transformer to achieve a similar performance with NAT. The comparison is based on the assumption that a smaller AT will suffer from more information loss than a larger AT. We use Transformer-\textit{base} for NATs, and the AT architecture is choose from Table \ref{tab:parameter_hyper}.

As shown in Table.\ref{tab:parameter_comparison}, we find that an AT only requires about 2.1\% $\sim$ 3.2\% of parameters to achieve similar performance with the NAT on WMT14 En-De. However, on Synthetic A, an AT requires at least 27.3\% of parameters to compete with the NAT. The results verify that large $\tc$ brings much information loss, making AT easily outperform NAT with much less parameters.

\begin{table}[!h]
\caption{Estimated $\tc$ and parameter size for an AT to achieve similar performance with NAT on various dataset. \textit{Parameter Ratio} is the ratio of AT parameter size and NAT parameter size. 
For example, ATs with 2.1\% $\sim$ 3.2\% of parameters achieve 4.60 $\sim$ 11.88 BLEU on WMT14 En-De, which is similar to the NAT performance (11.79).
The large $\tc$ leads to serious information loss in NAT learning, therefore a small AT can easily outperforms the NAT.}
\vspace{-0.5em}
\begin{center}
\resizebox{0.6\linewidth}{!}{
\setlength{\tabcolsep}{2mm}{
\begin{tabular}{c|cccc}
\hline
\bf Dataset         & \bf $\tc$ & \bf Parameter Ratio & \bf $\text{BLEU}_{\text{NAT}}$ & \bf $\text{BLEU}_{\text{AT}}$ \\
\hline
WMT14 En-De & 2.50 & 2.1\% $\sim$ 3.2\%  & 11.79 & 4.60 $\sim$ 11.88\\
WMT16 En-Ro & 2.20 & 3.2\% $\sim$ 4.3\% & 23.72 & 20.50 $\sim$ 24.75\\
Synthetic B & 1.51 & 4.3\% $\sim$ 7.0\% & 15.31 & 14.10 $\sim$ 16.10\\
Synthetic A & 0.92 & 27.3\% $\sim$ 100\% & 26.61 & 23.99 $\sim$ 26.96 \\
\hline
\end{tabular}
}
}
\end{center}
\label{tab:parameter_comparison}
\vspace{-0em}
\end{table}

\begin{table}[!h]
\caption{AT Architectures used in searching the parameter size.}
\vspace{-1em}
\begin{center}
\begin{small}
\setlength{\tabcolsep}{2mm}{
\begin{tabular}{c|cccccc}
\hline
     & \bf $d_{model}$ & \bf $d_{hidden}$ & \bf $n_{layers}$ & \bf $n_{heads}$ & \bf \# Param & \bf Parameter Ratio\\
\hline
1 & 32 & 128 & 2 & 2 & 1.3M & 2.1\% \\
2 & 48 & 192 & 2 & 2 & 2.0M & 3.2\% \\
3 & 64 & 256 & 2 & 2 & 2.7M & 4.3\% \\
4 & 96 & 384 & 2 & 2 & 4.4M & 7.0\% \\
5 & 128 & 512 & 3 & 4 & 6.6M & 10.5\% \\
6 & 256 & 1024 & 3 & 4 & 17.1M & 27.3\% \\
7 & 512 & 2048 & 6 & 8 & 62.6M & 100\% \\
\hline
\end{tabular}
}
\end{small}
\end{center}
\label{tab:parameter_hyper}
\end{table}

\section{Results on WMT17 Zh-En}
\label{app:results_on_zhen}

We repeat the experiments in Sec.\ref{sec:analysis_target} and Sec.\ref{sec:analysis_latent_input} on WMT17 Zh-En. As shown in Table \ref{tab:zhen_compare_target} and Table \ref{tab:zhen_compare_input}, our objective is strongly correlated with the translation quality, which supports our claim well.

\begin{table}[!h]
\caption{Comparison of methods that obtain proxy targets on WMT17 Zh-En. All methods use Vanilla and $\Linput$=0. $\hatLMPLE$ and BLEU are strongly correlated (Pearson's $|r|$=0.96). AXE's $\tau$ and OaXE's numbers indicate the skip penalty and the pre-training step, which are hyper-parameters in choosing proxy targets.}
\vspace{-0.5em}
\begin{center}
\setlength{\tabcolsep}{1.0mm}{
\begin{tabular}{l|cccc}
\hline
\bf Models & \bf $\LNAT$ & \bf $\Lhattarget$ & \bf $\hatLMPLE$ & \bf BLEU \\
\hline
Raw Data & 4.43 & \bf -6.25 & -1.82 & 8.69 \\
KD & 2.85 & -5.72 & -2.87 & 15.53 \\
\hline
+ AXE($\tau$=1) & \bf 1.02 & -2.95 & -1.93 & 9.68 \\
+ AXE($\tau$=5) & 1.93 & -5.00 & -3.07 & 18.39 \\
+ AXE($\tau$=10) & 2.31 & -5.20 & -2.90 & 18.25 \\
\hline
+ OaXE(10k) & 1.46 & -3.38 & -1.92 & 12.31 \\
+ OaXE(50k) & 1.19 & -4.50 & -3.31 & 18.79 \\
+ OaXE(300k) & 1.15 & -4.66 & \bf -3.50 & \bf 19.46 \\
\hline
\end{tabular}
}
\end{center}

\label{tab:zhen_compare_target}

\end{table}

\begin{table}[!h]
\caption{Comparison of methods and variants that obtain proxy inputs on WMT17 Zh-En. All methods use KD and $\Lhattarget=-5.72$. Sample and Default indicate the BLEU score in Input Sampling and Default Decoding. $\hatLMPLE$ is strongly correlated with Sample BLEU (Pearson's $|r|$=0.99) but less correlated with Default BLEU ($|r|$=0.35).}
\vspace{-0.5em}
\begin{center}
\resizebox{0.5\linewidth}{!}{
\setlength{\tabcolsep}{1.0mm}{
\begin{tabular}{l|ccccc}
\hline
\bf Models & \bf $\Linput$ & \bf $\LNAT$ & \bf $\hatLMPLE$ & \bf Sample & \bf Default \\
\hline
Vanilla & \bf 0 & 2.85 & -2.87 & 15.53 & 15.53\\
\hline
CMLM & 1.13 & 0.76 & -3.83 & 19.74 & 14.12\\
+ Fixed & 0.29 & 1.13 & \bf -4.30 & \bf 20.81 & 14.49\\
\hline
GLAT & 0.33 & 1.26 & -4.13 & 20.73 & 22.51\\
+ Levenshtein & 1.08 & \bf 0.44 & -4.19 & 20.71 & 21.70\\
+ $P_{\text{ref}}$ & 0.40 & 1.60 & -3.73 & 18.98 & 21.22\\
+ $1-P_{\text{ref}}$ & 0.73 & 0.81 & -4.17 & 20.79 & \bf 21.51\\

\hline
\end{tabular}
}
}
\end{center}
\label{tab:zhen_compare_input}

\end{table}

\section{Details and Full Results of Dynamic KD}
\label{app:result_dynamickd}

\noindent \textbf{Decoding Tricks} \ \ 
In Table \ref{tab:dynamic_kd}, we apply some decoding tricks for the results on the last row:
\vspace{-1em}

\begin{itemize}[leftmargin=1em]
    \setlength{\itemsep}{0ex}{
    \setlength{\parskip}{2px}{
    \item We use length parallel decoding (LPD, \citealp{imitation2019wei}). We use a candidate set of 3. Since all candidates can be generated simultaneously, LPD is still much fast in inference. It is worth noting that LPD is faster than NPD \cite{nat2018gu} since it does not need an external reranker.
    \item We use the de-duplication trick \cite{iterativerefinement2018lee}, i.e., removing the repeated tokens in generated sentences.
    \item We adjust the predicted length by a factor $\lambda$ \cite{axe2020ghazvininejad}. The factor is tuned on the validation set. We use $\lambda=1$ (i.e., the predicted length is not changed) for WMT14 En-De, and $\lambda=1.05$ for WMT17 Zh-En.
    }
    }
\end{itemize}
\vspace{-1em}

\noindent \textbf{Full Results}\ \ 
In Table \ref{tab:compare_iterative}, we compare Dynamic KD against strong baselines including non-iterative and iterative NATs. Moreover, we justify the necessity of the regularizer in Eq.\ref{eq:objective} by removing $\Lhattarget$ in choosing the proxy target (i.e., $T^* = \argmin_T \LNAT$) as an ablation study.

\begin{table}[!h]
\caption{Full results of comparing Dynamic KD against AT and previous NATs. The best results of Non-iterative NATs are bolded.
NPD \cite{nat2018gu} and LPD \cite{imitatenat2019wei} indicate reranking methods with the number of candidates. NPD are slower than LPD due to the use of an external AT reranker. The ablation of regularizer in Dynamic KD indicates removing $\Lhattarget$ in choosing the proxy target. \dag: Results reported by previous studies. Speed up of Imputer are re-evaluated in our implementation. \S: Use adaptive
iteration numbers. }
\begin{center}
\resizebox{0.7\linewidth}{!}{
\setlength{\tabcolsep}{1.5mm}{
\begin{tabular}{c|l|c|ccc}
\hline
& \bf Models & $\#$ Iters & \bf En-De & \bf Zh-En & \bf Speedup\\
\hline
AT & Transformer & L & 27.11 & 23.89 & 1.0x\\
\hline
\multirow{8}{*}{\makecell{Iterative \\ NATs}} & \multirow{2}{*}{CMLM\dag~{\scriptsize\cite{cmlm2019ghazvininejad}}} & 4 & 25.94 & 21.90 & 3.0x \\
& & 10 & 27.03 & 23.21 & 1.3x \\
\cdashline{2-6}
& \multirow{3}{*}{DisCo\dag~{\scriptsize\cite{disco2020kasai}}} & 4 & 25.83 & 22.42 & 4.3x \\
& & 10 & 27.06 & 23.68 & 3.2x \\
& & $\approx$ 4 \S & 27.34 & 23.83 & / \\
\cdashline{2-6}
& \multirow{3}{*}{Imputer\dag~{\scriptsize\cite{imputermt2020saharia}}} & 1 & 25.8 & / & 14.9x \\
& & 2 & 27.5 & / & 7.5x \\
& & 8 & 28.2 & / & 2.7x \\
\hline
\multirow{3}{*}{\makecell{Non-iterative \\ NATs}} & MLE & 1 & 11.79 & 8.69 & 15.3x \\
& GLAT (NPD=7)\dag~{\scriptsize\cite{glat2021qian}}& 1 & 26.55 & / & 7.9x \\
& OaXE (LPD=5)\dag~{\scriptsize\cite{oaxe2021du}}& 1 & 26.1 & 22.1 & 14.2x \\
\hline
\multirow{8}{*}{Ours} & Vanilla + KD & 1 & 20.98 & 15.53 & 15.3x\\
& \textsf{+} Dynamic KD w/o Regularizer & 1 & 20.51 & 18.10 & 15.3x\\
& \textsf{+} Dynamic KD & 1 & 22.82 & 18.29 & 15.3x\\
& \quad \textsf{+} LPD=3 + Decoding Tricks & 1 & 24.83 & 19.97 & 14.6x \\
\cdashline{2-6}
& GLAT + KD & 1 & 25.12 & 22.51 & 15.3x\\
& \textsf{+} Dynamic KD w/o Regularizer & 1 & 22.66 & 22.14 & 15.3x\\
& \textsf{+} Dynamic KD & 1 & 25.88 & 23.07 & 15.3x\\
& \quad \textsf{+} LPD=3 + Decoding Tricks & 1 & \bf{26.89} & \bf 24.42 & 14.6x\\
\hline
\end{tabular}
}
}
\end{center}

\label{tab:compare_iterative}

\end{table}

\section{Formalization of Existing Methods in MPLE}
\label{app:instances}

In the main paper, we briefly describe how existing methods obtain $Z$ and $T$. In this section, we present detailed formalization of these methods by describing the heuristic rules and their objectives in the framework of MPLE.
Specifically, we formalize each method in two steps:

\textbf{First}, \textit{we define the variational distribution following their heuristic rule.} 
Existing methods use heuristic rule to obtain $Z$ and $T$, which builds the variational distribution $Q(T, Z|X)$ used in the derivation of MPLE (Eq.\ref{eq:variational}). For all methods in our analysis, their variational distribution is defined by
\begin{align}
    Q(T, Z|X) := Q(T|X) Q(Z|T, X),
\end{align}
where $Q(T|X)$ is defined by the methods that obtain the proxy target (including Raw Data, KD, AXE, and OaXE), and $Q(Z|T, X)$ is defined by the methods that obtain the proxy input (including Vanilla, CMLM, and GLAT).

\textbf{Second}, \textit{we prove that their original objective is equivalent to minimizing $\LNAT$ of Eq.\ref{eq:L_nat}.} 
Notably, in M-step, $Q$ is unchanged when optimizing the model $\theta$ as discussed in Sec.\ref{sec:understanding}, so we only prove that their objective is equivalent to minimizing the NLL:
\begin{align}
    \LNAT = - \E_{Q(Z, T|X)} \left[ \log P_{\theta}(T|Z, X) \right] + \text{Constant} \label{eq:lnat_reformulate}
\end{align}
For some methods, the proof is trivial and thus omitted in the following sections.

\subsection{Raw Data}
Raw Data uses the original target sentence as the proxy target.
Formally, it defines $Q(T|X)$ as a one-point distribution that $Q(T=Y^*|X) = 1$, where $Y^*$ is the original target in the dataset. 

\subsection{Knowledge Distillation (KD, \citealp{nat2018gu})}

KD first trains an autoregressive model $P_{AR}$ on the raw data, and then uses beam search to obtain $T^* = \argmax_{Y} P_{AR}(Y|X)$.
Formally, $Q(T|X)$ is defined as a one-point distribution at $T^*$.

\subsection{Aligned Cross Entropy (AXE, \citealp{axe2020ghazvininejad})}
\label{app:axe}

\begin{figure*}[!t]
  \centering
  \includegraphics[width=0.85\linewidth]{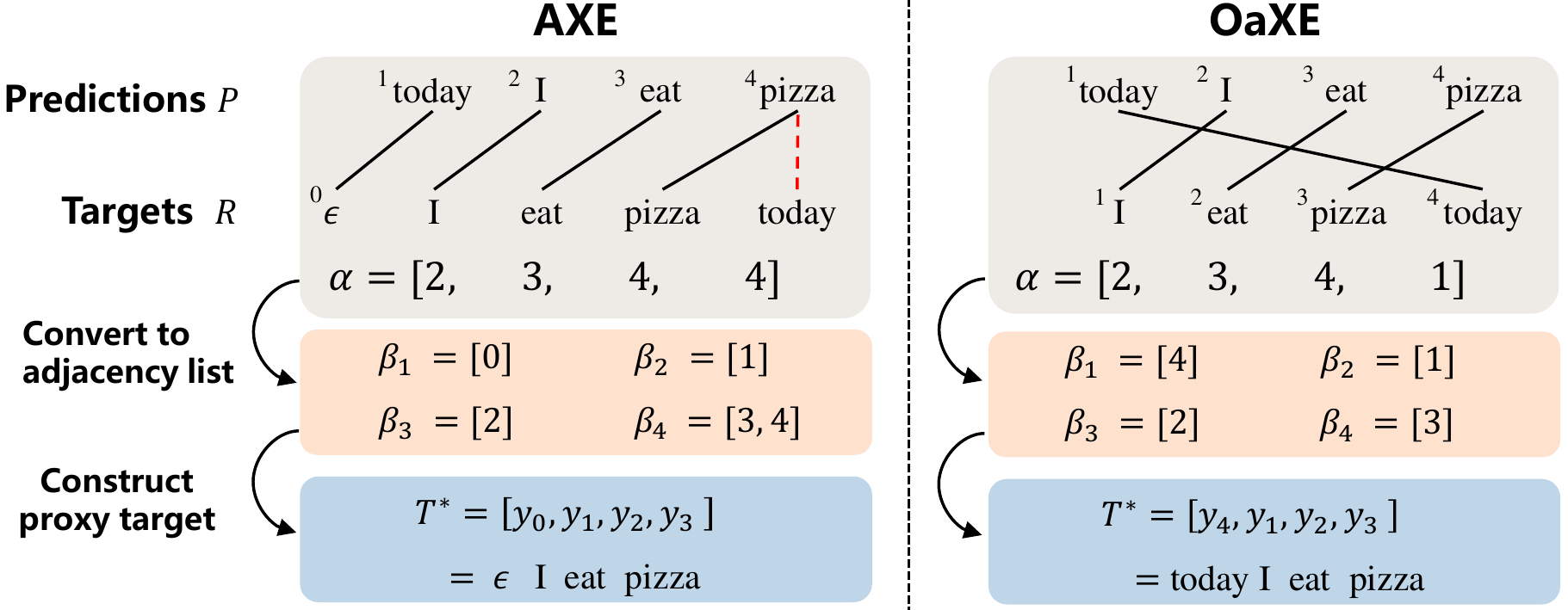}
  \caption{Examples of the alignment $\pmb{\alpha}$, the adjacency list $\pmb{\beta}$, and the proxy target $T^*$ in AXE and OaXE. Red dotted line in AXE indicates the \textit{skip target} operation.}
  \label{fig:axe-details}
  \vspace{-1em}
\end{figure*}

$Q(T|X)$ is defined as a one-point distribution at $T^*$, where $T^* = \argmin_{T \in \mathcal{S}(R)} \LNAT$. The reference $R = [r_1, \cdots, r_L]$ is picked from Raw Data or KD. Any $T \in \mathcal{S}(R)$ is a subsequence of $R$ with empty tokens $\epsilon$ inserted.\footnote{The NAT model may learn to predict empty tokens, which will be removed after generation.} An example is shown in Fig.\ref{fig:axe-details}.

\smallvspace
\noindent \textbf{Original Objective}\ \ 
AXE introduces a monotonic alignment $\pmb{\alpha} = [ \alpha_1, \cdots, \alpha_L ]$, where the $i$-th token of the reference $R$ is aligned to the $\alpha_i$-th token of the NAT prediction. Formally, the AXE loss is defined as
\begin{gather}
    \LL_{\text{AXE}} = \min_{\pmb{\alpha}} \left[ - \sum_{i=1}^{L} \log P_{\alpha_i}(r_i) - \sum_{k \notin \alpha} \log P_k(\epsilon) \right], \notag
     \\
    \text{s.t.}\ \ 1 \leq \alpha_1 \leq \cdots \leq \alpha_L \leq L. \notag
\end{gather}
The first term is the cross entropy between aligned targets and predictions, and the second term is a penalty for unaligned predictions.

In the AXE loss, a single prediction may be aligned to multiple target tokens. In their original paper, aligning the prediction to the first target token is called the ``align'' operation, and aligning the prediction to later tokens is called the ``skip target'' operation. However, a one-to-many alignment will damage the performance, so they penalize the ``skip target'' operations with a factor $\delta$. This trick is called the \textit{skip penalty}.

\smallvspace
\noindent \textbf{Proof of Equivalence}\ \ 
To connect their definition with ours, we convert the alignment to an adjacency list, as shown in Figure \ref{fig:axe-details}, where $\pmb{\beta}_i$ is a list containing all aligned tokens for the $i$-th prediction. Specially, if the $i$-th prediction is not aligned, we set $\pmb{\beta}_i = [0]$ and $r_0 = \epsilon$. Then, $\LL_{\text{AXE}}$ can be reformulated as
\begin{align}
    \min_{\pmb{\beta}} \left[ - \sum_{i=1}^{L} \log P_{i}(r_{\beta_{i, 1}}) - \delta \sum_{i=1}^{L} \sum_{j=2}^{|\pmb{\beta}_i|} \log P_{i}(r_{\beta_{i, j}}) \right], \label{eq:axe_reformulate}
\end{align}
where $\beta_{i, j}$ indicates the $j$-th element of $\pmb{\beta}_i$. The first term is the cross entropy between the prediction and a new target $T^* = [ r_{\beta_{1, 1}}, \cdots, r_{\beta_{L, 1}} ]$, and the second term is the penalty for ``skipping target'' operations.

When $\delta = 0$, Eq.\ref{eq:axe_reformulate} is equivalent to finding an optimal $T^*$ to minimize $\LNAT$ in Eq.\ref{eq:lnat_reformulate}.
Since $\pmb{\alpha}$ is a monotonic alignment, $T^*$ is constrained and should be a subsequence of $R$ with some empty tokens inserted, which recover our definition.
When $\delta \neq 0$, the second term can be regarded as a regularizer to control the distortion between proxy targets and real targets.

\subsection{Order-agnostic Cross Entropy (OaXE, \citealp{oaxe2021du})}
\label{app:oaxe}

OaXE is similar to AXE despite the constraint $\mathcal{S}(R)$.
Any $T \in \mathcal{S}(R)$ is a permutation of $R$. An example is shown in Fig.\ref{fig:axe-details}.

\smallvspace
\noindent \textbf{Original Objective}\ \ 
Different from AXE, OaXE's $\pmb{\alpha}$ is a non-monotonic alignment, and each predicted token can only be used once. The loss is defined as
\begin{gather}
    \LL_{\text{OaXE}} = \min_{\pmb{\alpha} \in \text{Perm}(L)} \left[ - \sum_{i=1}^{L} \log P_{\alpha_i}(r_i) \right], \notag
\end{gather}
where $\text{Perm}(L)$ indicates the permutations of sequences containing $1$ to $L$. 

\smallvspace
\noindent \textbf{Proof of Equivalence}\ \ 
Similar to the derivation for AXE, we can reformulate $\LL_{\text{OaXE}}$ as
\begin{align}
    \min_{\pmb{\beta}} \left[ - \sum_{i=1}^{L} \log P_{i}(r_{\beta_{i, 1}}) \right]. \notag
\end{align}
The above formulation recovers our definition: It finds an optimal $T^*$ to minimize $\LNAT$ in Eq.\ref{eq:lnat_reformulate}, where $T^*$ can be an arbitrary permutation of $R$.

However, without the monotonic constraints, $T^*$ in OaXE may be heavily distorted from the real target $Y$. To alleviate the problem, OaXE first pretrains the NAT using the vanilla MLE and then finetunes it to minimize $\LL_{\text{OaXE}}$. This trick is based on an intuition that the optimal $T^*$ in a well-trained NAT will be close to the real target.

\subsection{Vanilla}

Many NATs use a full masked sequence as $Z$ or predict $Z$ by Uniform Copy~\cite{nat2018gu} or attention~\cite{glat2021qian}. We regard them as vanilla decoder inputs because they do not introduce any hints from the target. Formally, they can be formulated as a one-point distribution $Q(Z = Z^*|T, X) = 1$, where $Z^*$ is obtained from a deterministic function $f(X)$.

\subsection{CMLM \cite{cmlm2019ghazvininejad} / GLAT \cite{glat2021qian}}
\label{app:cmlm_glat}

CMLM and GLAT sample the proxy input by randomly masking the target sentence.
Specifically, they first sample $l \in [1, L]$ as the number of unmasked tokens, and then obtain the proxy input by randomly masking $L-l$ tokens.

In Table \ref{tab:compare_input}, we compare CMLM, GLAT and their variants. Here we list their differences:

\vspace{-0.5em}
\begin{itemize}
\setlength{\itemsep}{0ex}{
\setlength{\parskip}{2px}{
    \item \textbf{CMLM} sets the number of unmasked tokens $l=\lambda L$, where $\lambda$ is uniformly sampled from 0 to 1.
    \item \textbf{CMLM + fixed masking ratio} uses $l=0.2L$ instead of random sampling.
    \item \textbf{GLAT} uses an adaptive sampling strategy according to the NAT prediction accuracy. Specifically, $l = \lambda \sum_{i=1}^{L} [T_i \ne \argmax(t_i|X)]$. We follow their original paper and anneal $\lambda$ from 0.5 to 0.3.
    \item \textbf{GLAT + mask by $P_{\text{ref}}$} use the same $l$ as GLAT, but chooses the unmasked tokens according to the difficulties in predicting them, where the probability of an unmasked $z_i$ is proportional to the prediction probability $P_{\theta}(t_i|X)$.
    \item \textbf{GLAT + mask by $1-P_{\text{ref}}$} chooses the unmasked tokens proportional to  $1 - P_{\theta}(t_i|X)$.
    }}
\end{itemize}
\vspace{-0.5em}

\smallvspace
\noindent \textbf{Implementation Details of Input Predictor} \ \ 
In Eq.\ref{eq:input_predictor}, we mention that $Z$ is predicted non-autoregressively.
Concretely, $P_{\theta}(z_i|X)$ is composed of two modules: $P_{\theta}(z_i\text{ is masked}|X)$ predicts whether $z_i$ is a masked token, and $P_{\theta}(t_i|X)$ predicts the target token $t_i$ from the vocabulary if $z_i$ is not masked. Formally,
\begin{align}
    P_{\theta}(z_i|X) &= \left\{\begin{array}{ll}
        P_{\theta}(z_i\text{ is masked}|X), & \text{if } z_i \text{ is masked};\\
        (1 - P_{\theta}(z_i\text{ is masked}|X)) P_{\theta}(t_i|X),  & \text{if } z_i = t_i; \\
        0, & \text{otherwise}.
        \end{array} \right. \notag
\end{align}

Therefore, $\Linput$ can be formulated as
\begin{align}
    \Linput &= \E_{Q(Z|X)} \left[ -\sum_{i=1}^{L} \log P_{\theta}(z_i|X) + \log Q(Z|X) \right], \notag
\end{align}
where $Q(Z|X)$ can be obtained according to the definition of heuristic rules.

For the first module $P_{\theta}(z_i\text{ is masked}|X)$, we reuse the Transformer encoder and the NAT decoder and further add a binary classification layer on top of the NAT decoder.
For the second module $P_{\theta}(t_i|X)$, we use a pre-trained vanilla NAT and freeze its parameters during the training of CMLM or GLAT. In this way, $P_{\theta}(t_i|X)$ can be computed offline to speed up the training.

For Input Sampling used in Table \ref{tab:compare_input}, we only do sampling from $P_{\theta}(z_i\text{ is masked}|X)$. If $z_i$ is not masked, we directly use $z_i = \argmax_{t_i} P_{\theta}(t_i|X)$ because it empirically leads to better performance.

\smallvspace
\noindent \textbf{Original Objective} \ \ 
In the original implementation, CMLM and GLAT use a masked language model objective, where the unmasked tokens are not included in the loss $\LNAT$. Formally,
\begin{align}
    \LNAT = \E_{Q(Z, T|X)} \left[-\sum_{i \notin \mathcal{G}} \log P_{\theta}(t_i|Z, X) \right], \notag
\end{align}
\noindent where $\mathcal{G}$ is the set of the unmasked token.

\smallvspace
\noindent \textbf{Proof of Equivalence} \ \ 
To reach a same formulation of Eq.\ref{eq:lnat_reformulate}, we add a \textit{copy mechanism} in the NAT decoder. The decoder directly copies the unmasked token as the prediction if available. As a result, for an unmasked token $t_i$, $\log P_{\theta}(t_i|Z, X) = 0$ because the prediction of $t_i$ is always correct. Therefore, the masked language model objective recovers our objective:
\begin{align}
    \LNAT &= \E_{Q(Z, T|X)} \left[-\sum_{i \notin \mathcal{G}} \log P_{\theta}(t_i|Z, X) \right] + 0 \notag \\
    &= \E_{Q(Z, T|X)} \Big[-\sum_{i \notin \mathcal{G}} \log P_{\theta}(t_i|Z, X) - \sum_{i \in \mathcal{G}} \log P_{\theta}(t_i|Z, X) \Big] \notag \\
    &= \E_{Q(Z, T|X)} \left[ -\log P_{\theta}(T|Z, X) \right] \notag
\end{align}
Note that the \textit{copy mechanism} does not require modifications to the network architecture.

\subsection{VAE \cite{vae2020shu}}

Although not discussed in our main analysis, VAE and its variants \cite{discretelatent2018kaiser, CNAT2021bao, latentglat2022} can also be formulated as a method to provide proxy input in MPLE.
VAE uses two trainable networks, the prior and posterior networks, to model $P_{\theta}(Z|X)$ and $Q(Z|T, X)$, respectively.
Specially, the posterior network $Q(Z|T, X)$ can be trained together with $\theta$.

\section{Implementation Details of Dynamic KD}
\label{app:dynamic_kd}

\begin{table}[!h]
\caption{Hyper-parameters and performance of AT teachers, which generate the target candidates in Dynamic KD.}
\vspace{-1em}
\begin{center}
\setlength{\tabcolsep}{2mm}{
\begin{tabular}{c|cccc}
\hline
    \bf Model & \bf \textit{tiny} & \bf \textit{small} & \bf \textit{base} & \bf \textit{big} \\
\hline
$d_{model}$ & 128 & 256 &  512 & 1024 \\
$d_{hidden}$ & 512 & 1024 &  2048  & 4096 \\
$n_{layers}$ & 3 & 3  &  6 & 6 \\
$n_{heads}$ & 4 & 4 & 8 & 8 \\
Dropout & 0.1 & 0.1 & 0.3 & 0.3 \\
\hline
WMT14 En-De & 20.46 & 24.29 & 27.11 & 28.49 \\
WMT17 Zh-En & 19.38 & 22.47 & 23.89 & 24.84 \\
\hline
\end{tabular}
}
\end{center}

\label{tab:dynamic_kd_hyper}
\end{table}

\textbf{Candidate Generation.}\ \ \ Dynamic KD chooses the proxy target from a candidate set $\Gamma$, which contains Raw Data and four distilled targets.
We generate the distilled targets with beam size 5 from four AT teachers, whose hyper-parameters and performance are shown in Table \ref{tab:dynamic_kd_hyper}.
For WMT14 En-De, we train the AT teachers for 100k updates with a batch of approximately 64k tokens. For WMT17 Zh-En, we raise the step to 300k to match the size of training data, and tune the length penalty in the beam search on the validation set.

\textbf{Candidate Selection Criterion.}\ \ \ Dynamic KD chooses the proxy target $T$ by minimizing $\LNAT + \Lhattarget$. However, $\Lhattarget$ requires samples from $\Pdata(Y|X)$ as defined in Eq.\ref{eq:L_target_hat}, which is intractable on the training set.
To tackle the issue, we approximate $\Lhattarget$ by the pairwise BLEU between the candidates: \footnote{We assume $Q(T|X)$ is a one-point distribution on the selected proxy target $T$.}
\begin{align}
    \Lhattarget &= -\beta~\E_{\Pdata(Y|X)} \left[S(Y, T)\right] \notag \\
    &\approx -\beta \sum_{i=1}^{5} \gamma_{i} S(\Gamma_i , T) \triangleq \Lhattargetprime, \label{eq:multikd_target}
\end{align}
where $S$ is the sentence BLEU, $\Gamma_i$ is the target distilled from the $i$-th teacher model, and $\gamma_i$ is hyper-parameters to bias the candidates from different teachers ($i=5$ indicates Raw Data). We use $\Lhattargetprime$ instead of $\Lhattarget$ in selecting the proxy target.

\textbf{Hyper-parameter Selection.}\ \ \ To find the optimal value of $\gamma_i$, we introduce the multi-reference dataset \cite{uncertainty2018ott, parity2018hassan}, making it possible to adjust the value of $\Lhattargetprime$ according to the real $\Lhattarget$.
Intuitively, if $\LNAT$ is the same for all target candidates, we should choose a proxy target that minimize the data distortion.
Therefore, we obtain $T = \argmin \Lhattargetprime$ as the current proxy target with a specific $\gamma_i$, and then evaluate the real data distortion $\Lhattarget$. We tune $\gamma_i$ to minimize $\Lhattarget$.
Notably, tuning $\gamma_i$ only involves calculating the BLEU score, which does not need to train a NAT model. We do a manual search from 1 to 3 with the step of 0.1 and finally choose $\gamma = [0.7, 0.6, 1.3, 1.5, 2.3]$ for WMT14 En-De and $\gamma = [0.9, 1.4, 1.3, 0.9, 2.0]$ for WMT17 Zh-En. Then we train a NAT with the dynamic KD and further tune $\beta$ according to the generation performance on the validation set, where we finally choose $\beta=0.2$ for WMT14 En-De and $\beta=0.1$ for WMT17 Zh-En.

\textbf{Applying Dynamic KD to GLAT.}\ \ \ When combining GLAT with Dynamic KD, $\LNAT$ may suffer from high variance because $Z$ is sampled from $Q(Z|T)$ following the rule of GLAT.
In our implementation, we simply ignore $Z$ when choosing the proxy target. Specifically, we obtain the proxy target by $T^* = \argmin_T \hat{\LL}_{\text{NAT}} + \Lhattargetprime$, where
\begin{align}
    \hat{\LL}_{\text{NAT}} &= \E_{Q(T|X)} \left[ -\log P_{\theta}(T|Z^*, X) \right], \notag
\end{align}
\noindent and $Z^*$ is a full masked sequence.


\section{Details of Experiment Settings}
\label{sec:experiment-details}

For WMT14 En-De, we follow \citet{kdnat2020zhou} to use a joint BPE \cite{bpe2016sennrich} with 32K merge operations, which leads to a vocabulary of 40k tokens.
For WMT17 Zh-En, we follow \citet{disco2020kasai} to use a BPE with 32K merge operations, which leads to vocabularies of 48k tokens in Chinese and 33k tokens in English.

All our models are implemented with Fairseq \cite{fairseq2019ott} and generally follow the hyper-parameter of transformer-\textit{base} \cite{transformer2017vaswani}. For regularization, we set dropout to 0.1, weight decay to 0.01, and label smoothing to 0.1.
Except for OaXE, all models are trained for 300k updates with a batch of approximately 64k tokens. The learning rate warms up to $5 \cdot 10^{-4}$ within 10k steps and then decays with the inverse square-root schedule.
For OaXE, we choose a pre-trained vanilla NAT and finetune the model for 100k steps with a fixed learning rate of $10^{-5}$.
We evaluate the BLEU scores on the validation set every epoch and average the best 5 checkpoints for the final model. All models are trained with mixed precision floating point arithmetic on 8 Nvidia V100-32G GPUs. It costs approximately 20 hours for a vanilla NAT and 30 hours for Dynamic KD + GLAT.

For fair comparisons in Table \ref{tab:compare_target} and \ref{tab:compare_input}, we do not use any decoding tricks and only modify the methods for obtaining $Z$ and $T$. Taking OaXE as an example, our implementation differs from their original paper \cite{oaxe2021du} in: (1) Our OaXE is finetuned on a vanilla NAT, not a CMLM. (2) They use Transformer-\textit{big} for KD whereas we use Transformer-\textit{base}. (3) They use Length Parallel Decoding \cite{imitatenat2019wei} of beam 5 and the de-duplication trick \cite{iterativerefinement2018lee} for decoding. We do not use any reranking methods here. (4) We do not use the truncation trick because it is incompatible with $\LNAT$ in our formulation.

\end{document}